\documentclass[10pt,twocolumn,letterpaper]{article}

\usepackage[pagenumbers]{cvpr} 

\def\eqref#1{Eq.~(\ref{#1})}

\usepackage{bm}
\usepackage{multirow}
\usepackage[linesnumbered,ruled,vlined]{algorithm2e}
\usepackage[most]{tcolorbox}
\usepackage{cuted}
\usepackage{makecell}

\definecolor{cvprblue}{rgb}{0.21,0.49,0.74}
\usepackage[pagebackref,breaklinks,colorlinks,allcolors=cvprblue]{hyperref}

\def\paperID{44883} 
\def\confName{CVPR}
\def\confYear{2026}

\title{PromptLoop: Plug-and-Play Prompt Refinement \\ via Latent Feedback for Diffusion Model Alignment}

\author{Suhyeon Lee \qquad Jong Chul Ye\\
Kim Jaechul Graduate School of AI, KAIST \\
{\tt\small \{suhyeon.lee, jong.ye\}@kaist.ac.kr}
}

\begin{document}
\begin{strip}
\vspace{-3.75em}
\maketitle
\vspace{-2em}
\centering
\includegraphics[width=0.8\textwidth]{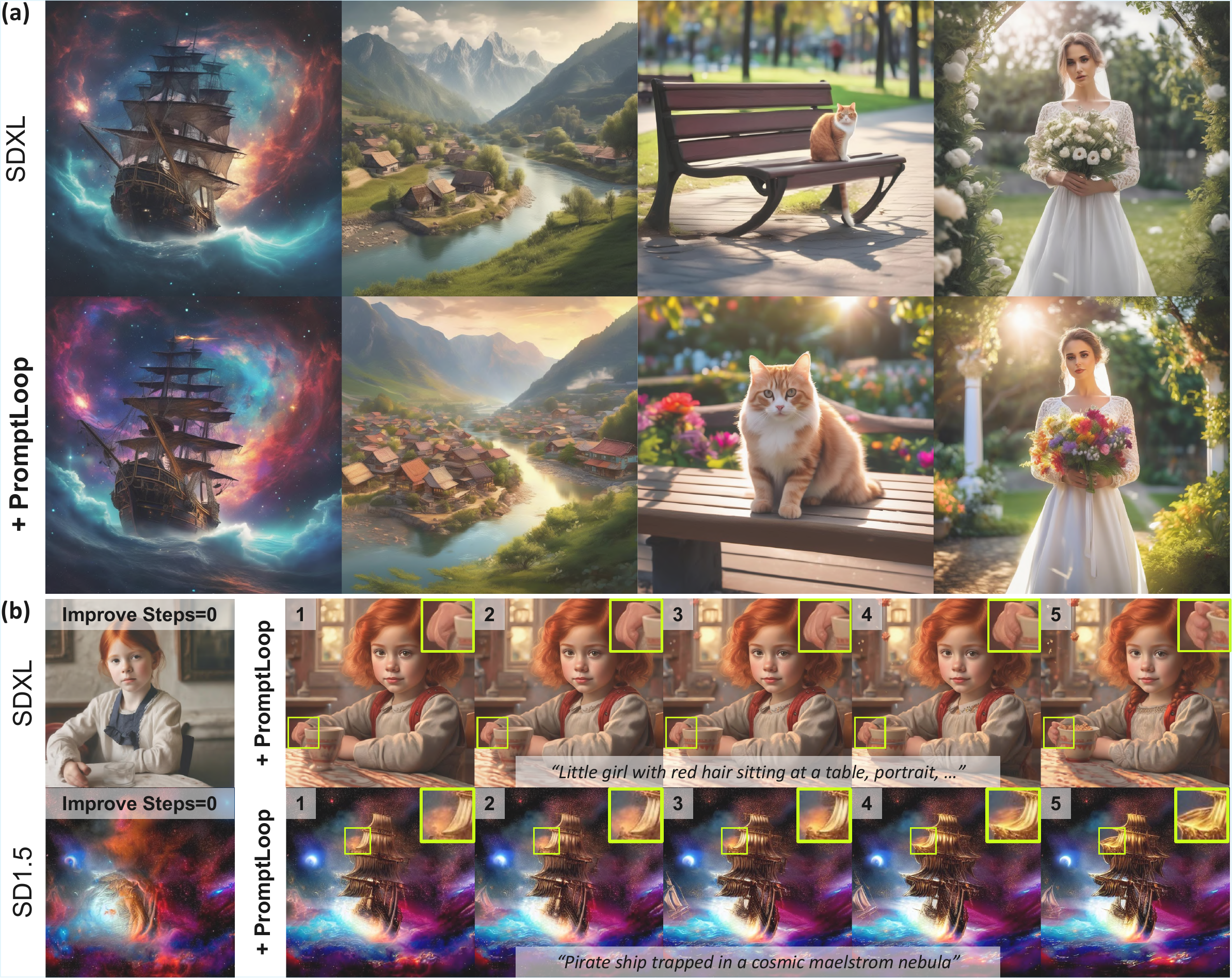}
\captionof{figure}{(a) PromptLoop uses latent feedback for stepwise prompt refinement, achieving functional equivalence to diffusion model RL and effective reward alignment (shown with ImageReward). (b) Multiple timestep-aware prompt updates during a single sampling yield stronger alignment.}
\label{fig:title}
\vspace{-0.5em}
\end{strip}

\begin{abstract}
Despite recent progress, reinforcement learning (RL)-based fine-tuning of diffusion models often struggles with generalization, composability, and robustness against reward hacking. Recent studies have explored prompt refinement as a modular alternative, but most adopt a feed-forward approach that applies a single refined prompt throughout the entire sampling trajectory, thereby failing to fully leverage the sequential nature of reinforcement learning. To address this, we introduce {\em PromptLoop}, a plug-and-play RL framework that incorporates latent feedback into step-wise prompt refinement. Rather than modifying diffusion model weights, a multimodal large language model (MLLM) is trained with RL to iteratively update prompts based on intermediate latent states of diffusion models. This design achieves a structural analogy to the Diffusion RL approach, while retaining the flexibility and generality of prompt-based alignment. Extensive experiments across diverse reward functions and diffusion backbones demonstrate that PromptLoop (i) achieves effective reward optimization, (ii) generalizes seamlessly to unseen models, (iii) composes orthogonally with existing alignment methods, and (iv) mitigates over-optimization and reward hacking while introducing only a practically negligible inference overhead.
\end{abstract}

\section{Introduction}
Diffusion models~\citep{ho2020denoising, song2020score, rombach2022high} have now become the state of the art for image generation.
Recently, increasing attention has been directed toward reinforcement learning (RL)-based approaches~\citep{sutton1998reinforcement} that align these models with user preferences through explicit reward optimization. Algorithms such as PPO~\citep{schulman2017proximal} and DPO~\citep{rafailov2023direct} have been applied directly to fine-tune diffusion model parameters~\citep{black2024training, wallace2024diffusion}. With reward functions defined over aesthetic quality, safety, human preference, or prompt alignment, these methods successfully steer model behavior without requiring new training data. However, direct RL fine-tuning remains limited: improvements often fail to generalize across models, additional enhancements are not easily composable once fine-tuning is complete, and pathological behaviors such as reward hacking or over-optimization can arise~\citep{kim2025test}.

In parallel, the rapid development of large language models (LLMs) \citep{brown2020language, grattafiori2024llama, guo2025deepseek} and multimodal large language models (MLLMs) \citep{liu2023visual, wang2024qwen2, wang2025internvl3} has inspired a new research direction: refining the input prompts rather than the diffusion model itself. These prompt-alignment methods either guide an LLM to improve a user’s prompt or adopt iterative feedback loops for prompt refinement~\citep{manas2024improving, yang2024idea2img, kim2025reward, khan2025test}. Building further, \cite{hao2023optimizing, wang2025promptenhancer, wu2025reprompt} propose to fine-tune LLMs with RL, enabling them to generate goal-directed prompt modifications more effectively. Compared to weight-level tuning, prompt refinement is attractive because prompts are shared across all text-to-image (T2I) models, inherently supporting generalization and orthogonal composability. Moreover, prompts, being abstract and discrete, may act as a buffer against reward hacking by decoupling reward optimization from direct parameter updates~\citep{lester-etal-2021-power, xie2022explanation, genewein2025understanding}. For a detailed discussion of related works, see Appendix~\ref{supp-related-works}. Nevertheless, prompt-based strategies remain structurally distinct from weight-level approaches. In diffusion models, parameters interact directly with intermediate latent variables $\bm{x}_t$ in a feedback loop, where each denoising step conditions on $\bm{x}_t$ to produce $\bm{x}_{t-1}$. By contrast, existing RL-based prompt refinement methods typically operate in a feed-forward manner, producing a refined prompt once and applying it uniformly across all timesteps, without leveraging the evolving latent trajectory.

To bridge this gap, we propose a generalized RL-based reward alignment framework called {\em PromptLoop} that achieves structural analogy to weight-level fine-tuning while preserving the modularity of prompt refinement (Fig.~\ref{fig:framework}). Specifically, our method introduces a plug-and-play prompt refinement module as a policy. This module leverages an MLLM to process feedback from the intermediate latent $\bm{x}_t$ as one of the states, analogous to diffusion RL formulations, and then refines the prompt $\bm{c}_t$ as the action injected into subsequent denoising steps. Thus, the sampling dynamics are adaptively adjusted without direct fine-tuning of the diffusion model itself. Unlike approaches that either delay feedback until after sampling or confine it to external loops, our method adopts a diffusion RL–style closed-loop design that embeds refinement directly within a single diffusion pass, ultimately enabling fine-grained adaptive control and improved efficiency. Despite the introduction of iterative MLLM-based policy inference, our implementation keeps the inference-time overhead at a practically manageable level compared to standard diffusion model sampling.

Extensive experiments across diverse diffusion models and reward functions demonstrate that our approach not only achieves effective reward optimization, but also generalizes robustly to unseen models, composes orthogonally with existing alignment methods, and mitigates over-optimization and reward hacking. These results establish PromptLoop as a practical and versatile approach to reward alignment for diffusion models. Our contributions are summarized as follows:

\begin{figure*}[t]
\begin{center}
\includegraphics[width=0.85\textwidth]{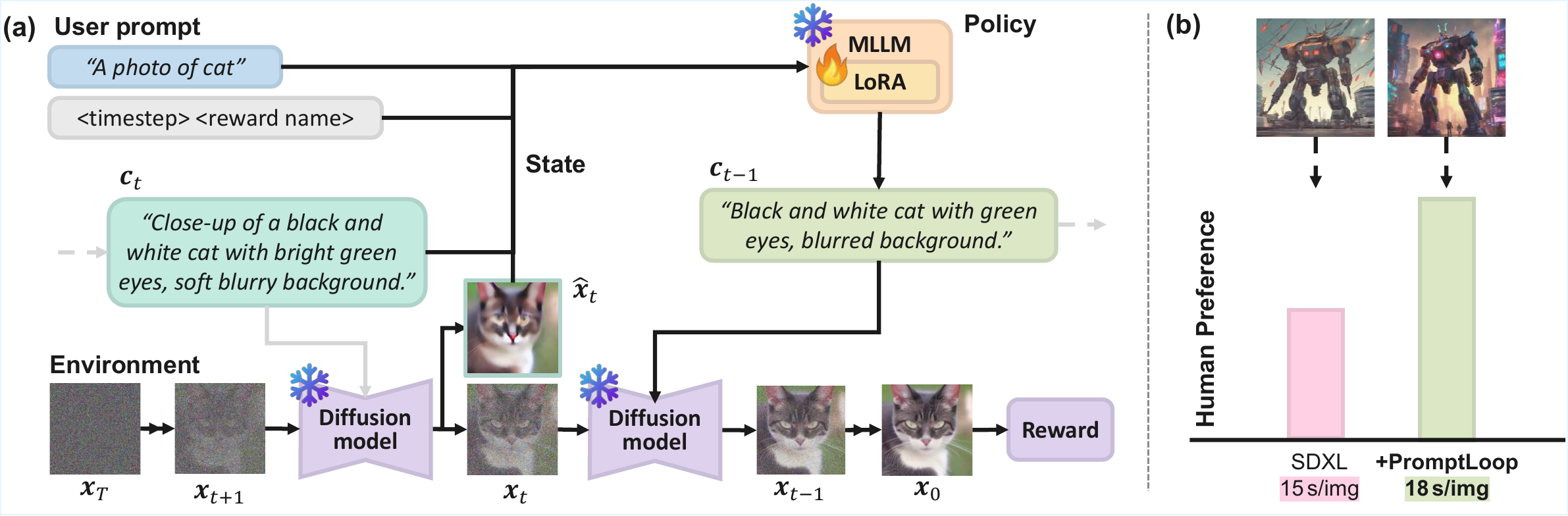}
\end{center}
\vspace{-1.0em}
\caption{\textbf{(a) Closed-loop prompt refinement framework with RL. (b) The proposed framework enhances human preference in a plug-and-play manner with minimal additional inference cost.} At each denoising step, the policy MLLM takes the current state—denoised estimates, the user query, and prior refinements—and generates an action, a refined prompt. The diffusion model then updates the state, and this loop continues until the final image is produced and scored by the reward model.}
\label{fig:framework}
\end{figure*}

\begin{itemize}
\item PromptLoop incorporates step-wise latent feedback into prompt refinement, achieving structural analogy to parameter-level tuning without modifying model weights.

\item We demonstrate broad generalization, effective reward optimization, and mitigation of reward hacking across diverse models and reward functions while incurring only minimal practical inference overhead.
\end{itemize}

\section{Preliminaries}

\textbf{Diffusion Models.} 
Diffusion models~\citep{ho2020denoising,song2019generative,sohl2015deep} are a class of latent variable generative models that approximate the data distribution $\bm{x}_0\sim p_\text{data}$ through a hierarchical latent process. The generative distribution is formulated as
\begin{equation}
    p_\phi(\bm{x}_0) = \int p(\bm{x}_T)\prod_{t=1}^T p^{(t)}_{\phi}(\bm{x}_{t-1}|\bm{x}_{t}) \, d\bm{x}_{1:T},
\end{equation}
where the prior $p(\bm{x}_T)$ is typically a standard Gaussian distribution. The latent sequence $\{\bm{x}_t\}_{t=1}^T$ is obtained via a forward noising process, which follows a Markov chain with a variance schedule $\{\beta_t\}_{t=1}^T$:
\begin{align}
    q(\bm{x}_t|\bm{x}_{t-1}) &= \mathcal{N}(\bm{x}_t \mid \sqrt{\alpha_t}\bm{x}_{t-1}, (1-\alpha_t)I), \\
    q(\bm{x}_t|\bm{x}_0) &= \mathcal{N}(\bm{x}_t \mid \sqrt{\bar{\alpha}_t}\bm{x}_0, (1-\bar{\alpha}_t)I),
\end{align}
where $\alpha_t = 1-\beta_t$ and $\bar{\alpha}_t = \prod_{i=1}^t \alpha_i$.
Training is carried out by learning to predict the injected Gaussian noise $\bm{\epsilon}$ using a neural network $\hat{\bm{\epsilon}}_\phi$, which is often conditioned by $\bm{c}$, known as \emph{$\epsilon$-matching}. This is equivalent to denoising score matching (DSM)~\citep{vincent2011connection, song2019generative}, which estimates the score function $\nabla_{\bm{x}_t}\log p(\bm{x}_t)$:
\begin{equation}
    \mathcal{L}_{\epsilon-\text{matching}} = 
    \mathbb{E}_{t,\bm{x}_0,\bm{\epsilon}} 
    \Big[ 
    \left\| \hat{\bm{\epsilon}}_\phi(\bm{x}_t,t,\bm{c}) - \bm{\epsilon} \right\|^2_2
    \Big],
\end{equation}
where $\bm{x}_t = \sqrt{\bar{\alpha}_t}\bm{x}_0 + \sqrt{1-\bar{\alpha}_t}\,\bm{\epsilon}$ with $\bm{\epsilon}\sim \mathcal{N}(0,I)$.
Once trained, the model iteratively reverses the noising process as follows:
\begin{align}
\bm{x}_{t-1} &= f(\bm{x}_t, \bm{z}_t, \bm{c}, t) \notag \\ &:=
\frac{1}{\sqrt{\alpha_t}}
\left(
\bm{x}_t - \frac{1-\alpha_t}{\sqrt{1-\bar{\alpha}_t}}
\hat{\bm{\epsilon}}_\phi(\bm{x}_t, t, \bm{c})
\right)
+ \sigma_t \bm{z}_t,
\end{align}
where  $\bm{z}_t \sim \mathcal{N}(0,I)$ and $\sigma_t^2 = \tfrac{1-\bar{\alpha}_{t-1}}{1-\bar{\alpha}_t}\beta_t$.   This corresponds to the canonical DDPM sampler~\citep{ho2020denoising}.
In general, $f(
\cdot)$ can be replaced by a variety of alternative samplers such as DDIM~\citep{song2020denoising}, PNDM~\citep{liu2022pseudo}, Euler~\citep{karras2022elucidating}, DPM-solver~\citep{lu2022dpm}.

\section{PromptLoop}

\begin{table}[!t]
\begin{minipage}[t]{1.00\linewidth}
    \centering
    \includegraphics[width=0.9\linewidth]{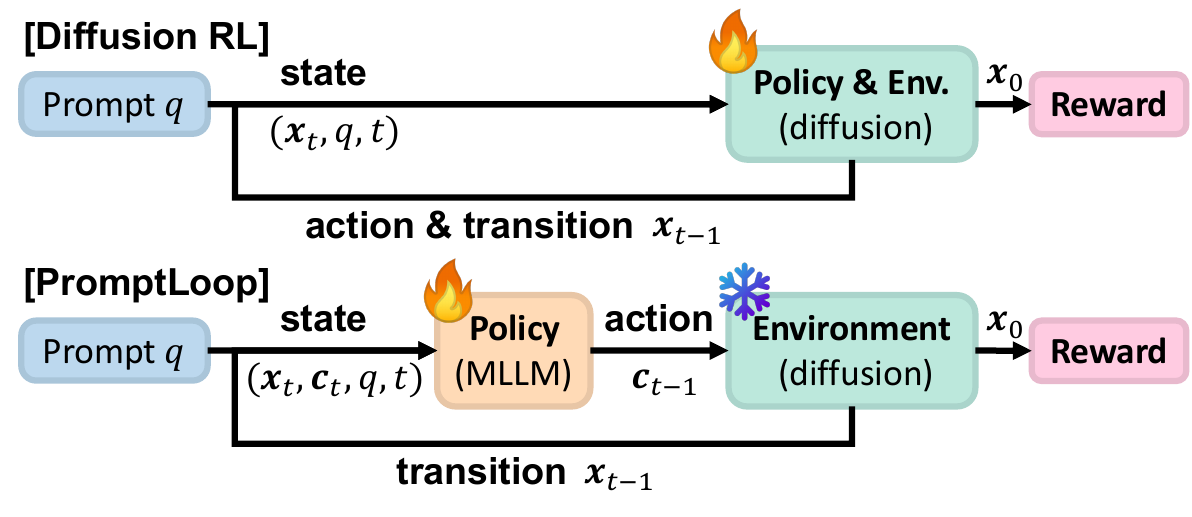}
    \vspace{0.5em}
\end{minipage} 
\begin{minipage}[t]{1.00\linewidth}
    \centering
    \resizebox{0.8\textwidth}{!}{

    \begin{tabular}{lcc}
    \toprule
     &  \textbf{Diffusion RL} & \textbf{PromptLoop (Ours)} \\ \midrule
    State $s_t$ & $(\bm{x}_t,q,t)$ & $(\bm{x}_t,\bm{c}_t,q,t)$ \\
    Policy & Diffusion model $p_\phi$ & MLLM $\pi_\theta$ \\
    Action $a_t$ & $\bm{x}_{t-1}\sim p_\phi(\cdot | s_t)$  & $\bm{c}_{t-1} \sim \pi_\theta (\cdot |s_t)$ \\
    Transition & -- & $\bm{x}_{t-1} =  f(\bm{x}_t, \bm{z}_t, \bm{c}_{t-1}, t)$\\
    Reward $R$ & $r(\bm{x}_0, q)$ & $r(\bm{x}_0, q)$ \\
    \bottomrule
    \end{tabular}
    }
    \captionof{figure}{\textbf{Structural analogy and key differences in MDP formulation between Diffusion RL and proposed framework.} Latent feedback establishes a functional correspondence with Diffusion RL, while PromptLoop diverges by adjusting the diffusion dynamics through timestep-aware prompts as actions.}
    \vspace{-1.0em}
    \label{fig:mdp}
  \end{minipage}
\end{table}

\subsection{MDP Formulation}
In PromptLoop, as shown in Fig.~\ref{fig:framework}, we aim to generate a refined text prompt $\bm{c}_{t-1}$ conditioned on user input $q$ and interpret intermediate visual states  $\bm{x}_{t}$ arising during the reverse diffusion process. The refined text prompt is then used to generate the next visual sample $\bm{x}_{t-1}$. To this end, we adopt a multimodal language model (MLLM)~\citep{liu2023visual, wang2024qwen2, wang2025internvl3} that accepts multimodal inputs and outputs refined prompts at each timestep. Then, our RL framework aims to train the MLLM to maximize the reward at the final visual
state $\bm{x}_{0}$. 
Formally, our Markov decision process (MDP) is defined as
the \(T\)-step reverse process with
 state $s_t$ and actions  $a_t$:
\begin{align}
  s_t = (\bm{x}_t, \bm{c}_t,q,t), \quad a_t = \bm{c}_{t-1},
  \end{align}
which are conditioned on an initial user prompt $q$, a previously updated prompt $\bm{c}_t$, and a visual state $\bm{x}_t$. Then, an action is sampled from the {\em MLLM policy} as $a_t \sim \pi_\theta(\cdot \mid s_t)$
and the visual state $\bm{x}_{t-1}$ is updated using the frozen diffusion model with the updated prompt $\bm{c}_{t-1}$.
A terminal reward \(r(\bm{x}_0,q)\) is assigned
at the final step.

This is in contrast to directly training the diffusion model's parameters~\citep{black2024training, wallace2024diffusion} using RL, where the MDP is defined with the state and action:
\begin{align}
  s_t = (\bm{x}_t, q, t), \quad a_t = \bm{x}_{t-1}
  \end{align}
where an action is sampled from the {\em diffusion} policy $\bm{x}_{t-1}\sim p_\phi(\cdot|s_t)$. The difference between the original Diffusion-RL and our RL framework is detailed in Fig.~\ref{fig:mdp}.

Note that our MDP formulation provides a structural correspondence between diffusion-model-based RL and the prompt refinement framework, enabled by a timestep-aware closed-loop latent feedback mechanism. On the other hand, in direct fine-tuning of diffusion models using RL, the diffusion model should be trained as the optimization target. This direct RL fine-tuning remains limited: improvements often fail to generalize across models, additional enhancements are not easily composable once fine-tuning is complete, and pathological behaviors such as reward hacking or over-optimization can arise. In our framework, the timestep-aware prompt-level actions can approximate the functional role of weight-level control, while retaining plug-and-play modularity, generalization, composability, and robustness against reward hacking.

Furthermore,  our approach has fundamental advantages over other prompt fine-tuning approaches. Specifically, prior prompt-tuning approaches either lack an intrinsic feedback loop~\citep{hao2023optimizing, wu2025reprompt, wang2025promptenhancer} or deliver feedback only after full sampling~\citep{manas2024improving, kim2025reward, khan2025test}, making them fundamentally different from our MDP formulation.

\subsection{Optimization}

At the end of each episode (\ie, $\bm{x}_T,\bm{x}_{T-1},\dots,\bm{x}_0$), the fully generated image $\bm{x}_0$ is evaluated by a reward function $r$ to produce a reward $R=r(\bm{x}_0, q)$. This can encode diverse criteria such as aesthetic quality~\citep{schuhmann_improved_aesthetic_predictor_2025}, safety~\citep{laion_safety}, prompt alignment~\citep{radford2021learning}, or human preference~\citep{wu2023human, xu2023imagereward}. The diffusion model and the reward model are both treated as black-box components: no gradient flows through them, and the policy is updated solely based on observed rewards. 

Policy gradient methods~\citep{williams1992simple, sutton1999policy} optimize this objective by estimating gradients with respect to $\theta$. A widely used algorithm is Proximal Policy Optimization (PPO)~\citep{schulman2017proximal}, which improves stability by constraining policy updates through a clipped surrogate objective:
\begin{equation} \label{eq-ppo}
\begin{gathered}
    \mathcal{L}_{\text{PPO}}(\theta) =
    \mathbb{E}_t \left[
        \min \!\Big(
            \rho_t(\theta)\,\hat{A}_t,\;
            \mathrm{clip}(\rho_t(\theta), 1-\epsilon, 1+\epsilon)\,\hat{A}_t
        \Big)
    \right] \\ - \beta\, \mathrm{KL}\!\left[\pi_{\theta_{\text{old}}}(\cdot \mid s_t)\;\|\;\pi_\theta(\cdot \mid s_t)\right], \\
    \text{where} \quad 
    \rho_t(\theta) = 
    \frac{\pi_\theta(a_t \mid s_t)}{\pi_{\theta_{\text{old}}}(a_t \mid s_t)}.
\end{gathered}
\end{equation}
Here, $\beta$ is a hyperparameter controlling the strength of the KL penalty, and the advantage $\hat{A}_t$ measures how much better an action is than the expected value under the current policy. Specifically, Group Relative Policy Optimization (GRPO)~\citep{guo2025deepseek} replaces the advantage estimator with a group-normalized reward to stabilize training and reduce variance:
\begin{equation} \label{eq-grpo}
    A_i = \frac{r_i - \mathrm{mean}(\{r_j(\cdot)\}_{j=1}^G)}{\mathrm{std}(\{r_j(\cdot)\}_{j=1}^G)},
\end{equation}
where $\{r_j(\cdot)\}_{j=1}^G$ are the rewards of $G$ sampled outputs for the same prompt. Therefore, we employ the standard token-level GRPO~\citep{guo2025deepseek}. Each training episode is initialized with user prompts drawn from a prompt-only dataset and proceeds via an online, on-policy reinforcement learning procedure.

\subsection{Implementation}

As part of our implementation, we design the MLLM’s input to be denoised latent representations rather than raw noisy states $\bm{x}_t$. Specifically, we convert the noisy visual latent state $\bm{x}_t$ into its denoised estimate $\hat{\bm{x}}_t$, which lies closer to the data manifold and thus provides a more semantically meaningful input to the policy model:
\begin{gather}
   \hat{\bm{x}}_t = \frac{1}{\sqrt{\bar{\alpha}_t}} 
  \big(\bm{x}_{t+1} - \sqrt{1-\bar{\alpha}_t}\,\hat{\bm{\epsilon}}_\phi(\bm{x}_{t+1},\bm{c}_{t},t)\big).
\end{gather}
Notably, many studies adopt these estimates for the final sample estimation~\citep{chung2022diffusion, yu2023freedom}, even for evaluation with vision-language models~\citep{kim2025free2guide, singhal2025general}. We further provide empirical evidence to support this design choice (Appendix~\ref{supp-tweedie}).

While our framework achieves structural equivalence, it introduces additional computational overhead: the policy model must be invoked during every denoising step of the diffusion process. This requirement also significantly increases memory costs, as both the diffusion model and the policy MLLM must be co-resident on the accelerator (\textit{e.g.} VRAM), or alternatively, incur large transfer times under offloading. Such constraints not only limit practical applicability but also complicate seamless integration of our approach into existing user-level diffusion-based image generation pipelines.  

To mitigate these issues, we adopt a sparse refinement strategy, where \textit{prompt refinement steps} are defined as a set of timesteps $\mathcal{R} \subseteq \{1, \dots, T\}$ with $|\mathcal{R}|=N_R$. The policy model is applied only at these steps rather than at every denoising step. For example, if the policy refines the prompt at timestep $t_1$ and the next refinement occurs at $t_2$ with $t_1 > t_2$, then $\bm{c}_{t_{1}-1:t_{2}} = \pi_\theta(\cdot \mid s_{t_1})$ and remains fixed until the next refinement step. During training, $\mathcal{R}$ is sampled uniformly at random, while during inference it is deterministically set at even intervals. This design allows the policy to generalize to an arbitrary number of refinement steps during sampling, while introducing only minimal inference overhead (see Sec.~Runtime Overhead).

We empirically observe that visual feedback from intermediate denoised states—though essential during training—is not strictly necessary at inference. Once the policy has learned the transition dynamics of the environment (\ie, the diffusion process coupled with the reward model), it can generate effective refinements without explicit access to intermediate visual signals. Consequently, refined prompts for all timesteps can be generated \textit{a priori}, allowing the diffusion process to proceed without interruptions during inference. This design yields substantial generalization capability and efficiency gains while remaining fully compatible with existing diffusion model ecosystems. It requires no modification to the generation loop and offers the ease of integration seen in feed-forward prompt optimization methods, yet uniquely retains the advantages of closed-loop RL fine-tuning.

\begin{figure*}
\begin{center}
\includegraphics[width=1.00\textwidth]{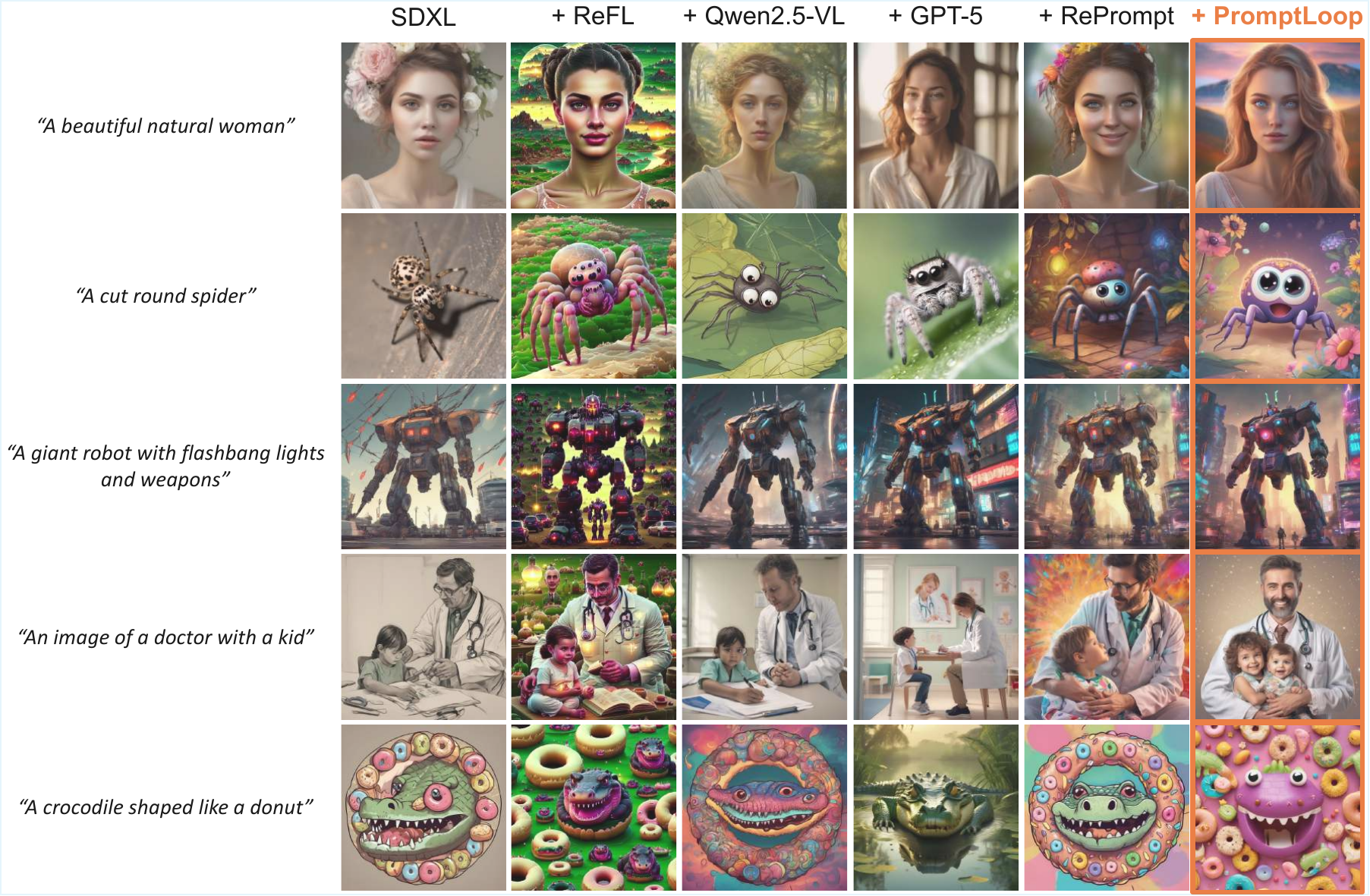}
\end{center}
\vspace{-1.0em}
\caption{Qualitative comparison of single-reward alignment, illustrating improvements over baseline methods. (SDXL \& ImageReward)}
\label{fig:result-ir-xl-compare}
\vspace{0.5em}
\end{figure*}

\begin{figure*}
\begin{center}
\includegraphics[width=1.00\textwidth]{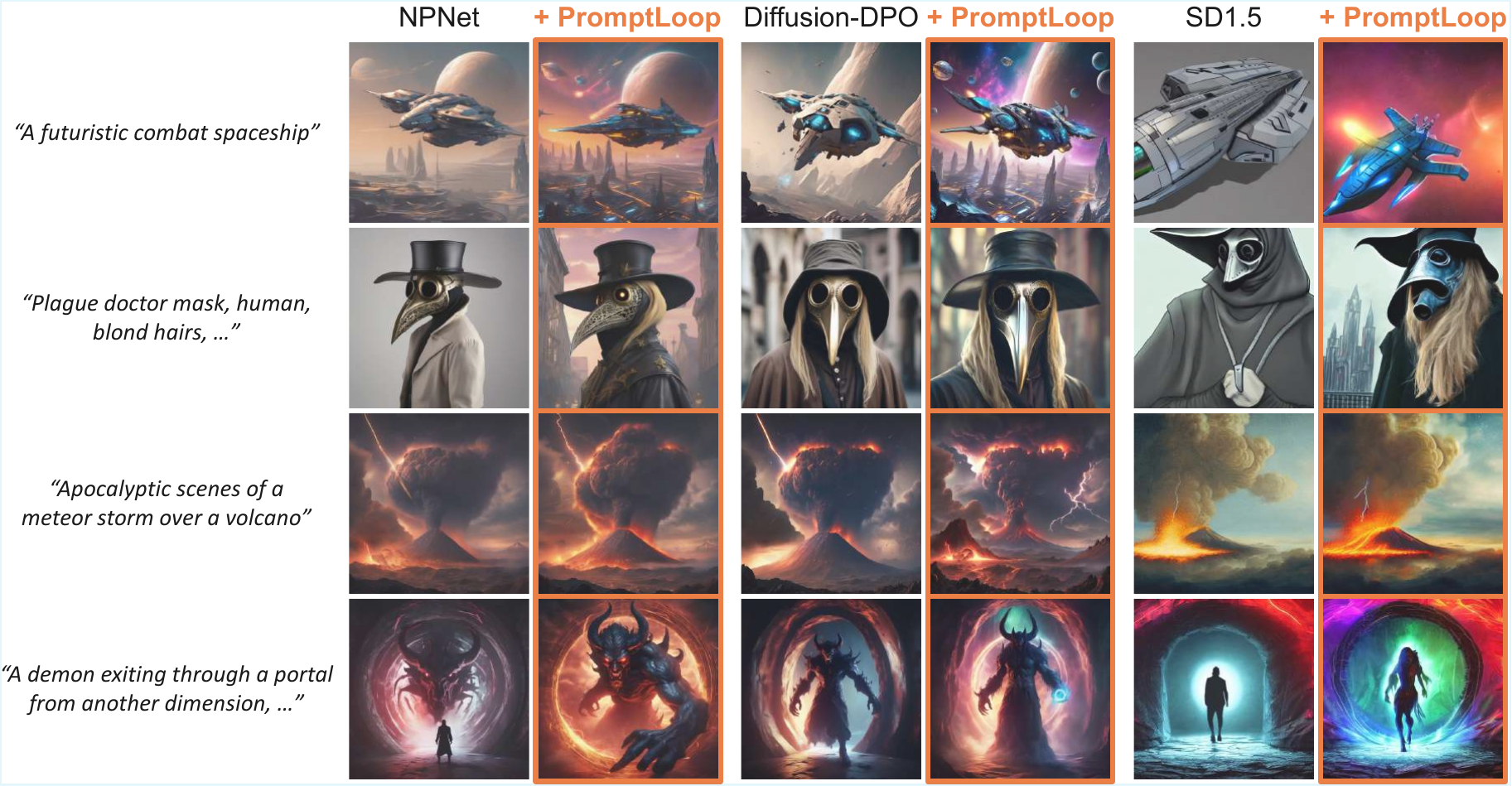}
\end{center}
\vspace{-1.5em}
\caption{Qualitative results showing the orthogonality and generalizability achieved by applying our method to unseen reward-alignment baselines (SDXL \& ImageReward).}
\label{fig:result-ir-xl-ortho}
\end{figure*}

\begin{table}
\caption{Quantitative evaluation on single-reward alignment with SD1.5 and SDXL, showing comparison with baselines and demonstrating orthogonality and generalizability.}
\label{tab:sigle-reward}
\centering
\resizebox{1.0\columnwidth}{!}
{
\newlength{\oldtabcolsep}
\setlength{\oldtabcolsep}{\tabcolsep}
\setlength{\tabcolsep}{2pt}
\small
\begin{tabular}{llcccc}
\toprule
\makecell{\textbf{Training} \\ \textbf{Setup}} & \textbf{Method}           & \makecell{\textbf{Image} \\ \textbf{Reward}} & \textbf{HPSv2} & \textbf{Aesthetics} & \makecell{\textbf{MLLM} \\ \textbf{Score}} \\
\toprule

 \multirow{13}{*}{\shortstack{SDXL \\ \& Image \\ Reward}} & SDXL & 0.7244 & 0.2805 & 6.073 & 0.735\\
 & + ReFL~\citep{xu2023imagereward} & 1.0119 & 0.2740 & 6.286 & 0.715  \\
 & + Qwen2.5-VL-3B~\citep{bai2025qwen2} & 0.5114 & 0.2739 & 6.279 & 0.741 \\
 & + GPT-5 & 0.6299 & 0.2794 & 6.231 & 0.740  \\
 & + RePrompt~\citep{wu2025reprompt} & 1.0148 & 0.2796 & 6.518 & 0.763 \\
 & \textbf{+ PromptLoop (ours)} & \textbf{1.0948} & \textbf{0.2807} & \textbf{6.583} & \textbf{0.764} \\ \cmidrule[\heavyrulewidth](l{0.5em}r{0.5em}){2-6}
 & SDXL + Diffusion-DPO~\citep{wallace2024diffusion} & 0.9921 & \textbf{0.2868} & 6.015 & 0.731 \\
 & \textbf{+ PromptLoop (ours)} & \textbf{1.2898} & 0.2862 & \textbf{6.491} & \textbf{0.763} \\ \cmidrule(l){2-6}
& SDXL + NPNet~\citep{zhou2025golden} & 0.7357 & 0.2805 & 6.059 & 0.733 \\ 
 & \textbf{+ PromptLoop (ours)} & \textbf{1.1213} & \textbf{0.2811} & \textbf{6.561} & \textbf{0.762} 
 \\ \cmidrule[\heavyrulewidth](l{0.5em}r{0.5em}){2-6}
 & SD1.5~\citep{rombach2022high} & 0.0816 & 0.2678 & 5.458 & 0.675 \\ 
 & \textbf{+ PromptLoop (ours)} & \textbf{0.4546} & \textbf{0.2688} & \textbf{5.813} & \textbf{0.723} \\

 \toprule

 \multirow{17}{*}{\shortstack{SD1.5 \\ \& Image \\ Reward}}  & SD1.5            & 0.0816 & 0.2678 & 5.458 & 0.675  \\ 
 & + DDPO~\citep{black2024training} & 0.6051 & 0.2726 & 5.562 & 0.693 \\
 & + ReFL~\citep{xu2023imagereward} & 0.6248 & {\textbf{0.2748}} & 5.577 & 0.691 \\ 
 & + Qwen2.5-VL-3B~\citep{bai2025qwen2} & -0.1720 & 0.2628 & 5.668 & 0.693 \\
 & + GPT-5~\citep{openai_gpt5_doc_2025} & -0.0950 & 0.2647 & 5.726 & 0.700 \\
 & + RePrompt~\citep{wu2025reprompt} & 0.4344 & 0.2684 & 5.850 & 0.722 \\
 & \textbf{+ PromptLoop (ours)}       & \textbf{0.6320} & 0.2701 & \textbf{5.853} & \textbf{0.725}           \\ \cmidrule[\heavyrulewidth](l{0.5em}r{0.5em}){2-6}
 & SD1.5 + DDPO~\citep{black2024training}       & 0.6051 & 0.2726 & 5.562 & 0.693           \\
 & \textbf{+ PromptLoop (ours)}  & \textbf{0.9842} & \textbf{0.2742} & \textbf{5.926} & \textbf{0.726}           \\ \cmidrule(l){2-6}
 & SD1.5 + DanceGRPO~\citep{xue2025dancegrpo}  & 0.6156 & 0.2795 & 5.662 & 0.704  \\
 & \textbf{+ PromptLoop (ours)}  & \textbf{0.8950} & \textbf{0.2799} & \textbf{5.990} & \textbf{0.728}           \\ \cmidrule(l){2-6}
 & SD1.5 + Diffusion-DPO~\citep{wallace2024diffusion}        & 0.3012 & 0.2717 & 5.568 & 0.687 \\
 & \textbf{+ PromptLoop (ours)}   & \textbf{0.7920} & \textbf{0.2739} & \textbf{5.968} & \textbf{0.734}           \\ \cmidrule(l){2-6}
 & SD1.5 + ReFL~\citep{xu2023imagereward}  & 0.6248 & {{0.2748}} & 5.577 & 0.691  \\
 & \textbf{+ PromptLoop (ours)} & \textbf{0.9271} & \textbf{0.2751} & \textbf{5.877} & \textbf{0.724} \\ \cmidrule[\heavyrulewidth](l{0.5em}r{0.5em}){2-6}
 & SDXL~\citep{podell2023sdxl} & 0.7244 & 0.2805 & 6.073 & 0.735 \\
 & \textbf{+ PromptLoop (ours)} & \textbf{1.0859} & \textbf{0.2807} & \textbf{6.535} & \textbf{ 0.763} \\
 
\bottomrule
\end{tabular}
\setlength{\tabcolsep}{\oldtabcolsep}
}
\end{table}

\begin{table}
\caption{Quantitative evaluation on composite-reward alignment with SDXL-turbo, showing comparison with baselines and demonstrating orthogonality and generalizability.}
\label{tab:complex-reward}
\centering
\resizebox{0.85\columnwidth}{!}
{
\setlength{\tabcolsep}{3pt}
\small
\begin{tabular}{llcccccc}
\toprule
\makecell{\textbf{Training} \\ \textbf{Setup} } & \textbf{Method}           & \textbf{GenEval} & \makecell{ \textbf{Image} \\ \textbf{Reward}} & \textbf{HPSv2} \\ 
\toprule
\multirow{8}{*}{\shortstack{SDXL-turbo \\ \& RePrompt}} 
 & SDXL-turbo~\citep{sauer2024adversarial} & 0.5445 & 0.7769 & 0.2915 \\
 & + Qwen2.5-VL-3B~\citep{bai2025qwen2} & 0.5212 & 0.6417 & 0.2893 \\
 & + GPT-5~\citep{openai_gpt5_doc_2025} & 0.5456 & 0.7593 & 0.2915  \\
 & + RePrompt~\citep{wu2025reprompt} & 0.5101 & 0.7876 & 0.2912 \\
 & \textbf{+ PromptLoop (ours)}  & \textbf{0.5483} & \textbf{0.8516} & \textbf{0.2938} \\
 \cmidrule[\heavyrulewidth](l{0.5em}r{0.5em}){2-5}
 & SDXL~\citep{podell2023sdxl}       & 0.5431 & 0.5518 & 0.2886  \\
 & \textbf{+ PromptLoop (ours)} &\textbf{ 0.5505} &  \textbf{0.7420} & \textbf{0.2906}  \\
 \cmidrule(l){2-5}
 & SD1.5~\citep{rombach2022high}        & 0.4206 & -0.1315 & 0.2783  \\
 & \textbf{+ PromptLoop (ours)} & \textbf{0.4399}  & \textbf{-0.0375}  & \textbf{0.2793}   \\
\bottomrule
\end{tabular}
\setlength{\tabcolsep}{\oldtabcolsep}
}
\end{table}

\begin{table}
\caption{Ablation study results showing the effectiveness of each proposed component.}
\label{tab:ablation}
\centering
\resizebox{0.95\columnwidth}{!}
{
\small
\begin{tabular}{lccc}
\toprule
\textbf{Components}  & \textbf{ImageReward} & \textbf{HPSv2} & \textbf{MLLM Score} \\
\toprule
 SD1.5 & 0.0816 & 0.2678 & 0.675 \\
 + policy model & -0.2315 & 0.2617 & 0.681 \\
 + GRPO training & 0.4344 & 0.2684 & 0.722 \\
 + multiple improvements & 0.4912 & 0.2690  & 0.724 \\
 + visual feedback & \textbf{0.6320} & \textbf{0.2701} & \textbf{0.725} \\
\bottomrule
\end{tabular}
}
\end{table}

\section{Experimental Results}

\subsection{Methods}

\textbf{Tasks.} 
To evaluate our framework as a general black-box reward alignment system, we consider two categories of reward models: \textit{single reward} and \textit{composite reward}. For the single reward setting, we adopt ImageReward~\citep{xu2023imagereward}, a widely used neural network–based reward function for human preference and prompt alignment, along with incompressibility, compressibility, and aesthetic score models~\citep{black2024training, schuhmann_improved_aesthetic_predictor_2025}. These rewards are applied to train Stable Diffusion v1.5~\citep{rombach2022high} (SD1.5) and Stable Diffusion XL~\citep{podell2023sdxl} (SDXL) using prompts from the Pick-a-Pic v2 dataset~\citep{kirstain2023pick}. For the composite reward setting, we follow a RePrompt-style design~\citep{wu2025reprompt}, which combines ImageReward, MLLM-reward~\citep{openai_gpt5mini_doc_2025}, and additional task-specific signals such as format and length reward. This composite reward style is intended to better capture human preference and object-focused alignment. Compared to the single reward setting, the composite reward is more complex and difficult to optimize, since it requires balancing multiple heterogeneous objectives simultaneously. We use it to train Stable Diffusion XL Turbo~\citep{sauer2024adversarial} (SDXL-turbo), a distillation model designed for few-step generation, with the prompt dataset introduced by~\citet{wu2025reprompt}.

\noindent \textbf{Evaluations.}
In evaluation, we validate our model’s capability along three aspects: performance, orthogonality, and generalizability. For performance evaluation, we compare against baseline reward alignment methods, including DDPO~\citep{black2024training}, ReFL~\citep{xu2023imagereward}, Qwen2.5-VL-3B~\citep{bai2025qwen2}, GPT-5~\citep{openai_gpt5_doc_2025}, and RePrompt~\citep{wu2025reprompt}. For orthogonality, we apply our trained policy model to other diffusion models that are fine-tuned or augmented with additional modules for human preference alignment, demonstrating that our method can be applied orthogonally to existing preference alignment techniques. Specifically, we evaluate on DDPO~\citep{black2024training}, DanceGRPO~\citep{xue2025dancegrpo}, Diffusion-DPO~\citep{wallace2024diffusion}, ReFL, and NPNet~\citep{zhou2025golden}. These experiments demonstrate that our method can be applied orthogonally to diverse alignment techniques without requiring retraining. For generalizability, we evaluate our trained policy model on different versions of text-to-image diffusion models that were not seen during training. It is important to note that for both orthogonality and generalizability, the policy model was only trained on the vanilla diffusion model environment, which differs from the sampling variants.

\subsection{Results}

\noindent \textbf{Single Reward.} After aligning SD1.5 and SDXL models with the ImageReward reward function, we conducted quantitative evaluations (Tab.~\ref{tab:sigle-reward}). The results demonstrate that our proposed methodology consistently outperforms baselines not only with respect to the target reward but also across most evaluation metrics.
Crucially, our method is orthogonal, demonstrating broad compatibility with a variety of human preference alignment strategies—including noise optimization, reinforcement learning, and gradient-based optimization—regardless of their internal mechanisms. Consequently, establishing superiority over full alignment frameworks is not the sole objective; rather, one of the key values of our approach is its ability to seamlessly enhance and extend these methods, not replace them.

The qualitative comparisons in Fig.~\ref{fig:result-ir-xl-compare},~\ref{fig:result-ir-xl-ortho}, which present SDXL results, highlight effective alignment to the reward signal, composability of our method, and robustness against over-optimization, an aspect not always captured by quantitative metrics. For instance, ReFL optimized the ImageReward signal through strategies resembling reward hacking from a human perspective. However, this degradation was not clearly reflected in commonly used metrics such as HPS or aesthetic scores. Thus, the qualitative evaluation further underscores the value of our approach in revealing such vulnerabilities.

\noindent \textbf{Composite Reward.} As one of the evaluation tasks, we consider RePrompt-style multi-reward alignment, which imposes challenging conditions such as a few-step distillation model and object-centric prompt alignment benchmarks (Tab.~\ref{tab:complex-reward}). Our framework achieves strong qualitative and quantitative results under these settings, showing consistently high performance across an object-centric prompt alignment benchmark and multiple human-preference benchmarks. This indicates that our method effectively avoids over-optimization while achieving robust alignment. Moreover, we observe similar generalization to diffusion models unseen during training.

\noindent \textbf{Ablation Studies.} We conducted a series of ablation studies to validate the contributions of our proposed components and to analyze the effects of key hyperparameters. All experiments were performed on a single reward task (ImageReward) using the SD1.5 model. Tab.~\ref{tab:ablation} summarizes the results, where each major component was added incrementally to highlight its individual effect. First, simply applying the policy model to improve prompts without training (+ policy model) degraded performance, as the model could not fully capture the task despite the use of a system prompt. Training the policy model with GRPO (+ GRPO training) led to significant improvements across all metrics. Incorporating multiple prompt refinements within a single diffusion trajectory (+ multiple improvements, 5 steps) further boosted performance. Finally, introducing training-time visual feedback substantially increased the target reward without reducing other metrics, suggesting that it helps mitigate reward hacking (+ visual feedback).

\begin{figure*}
\begin{center}
\includegraphics[width=0.75\textwidth]{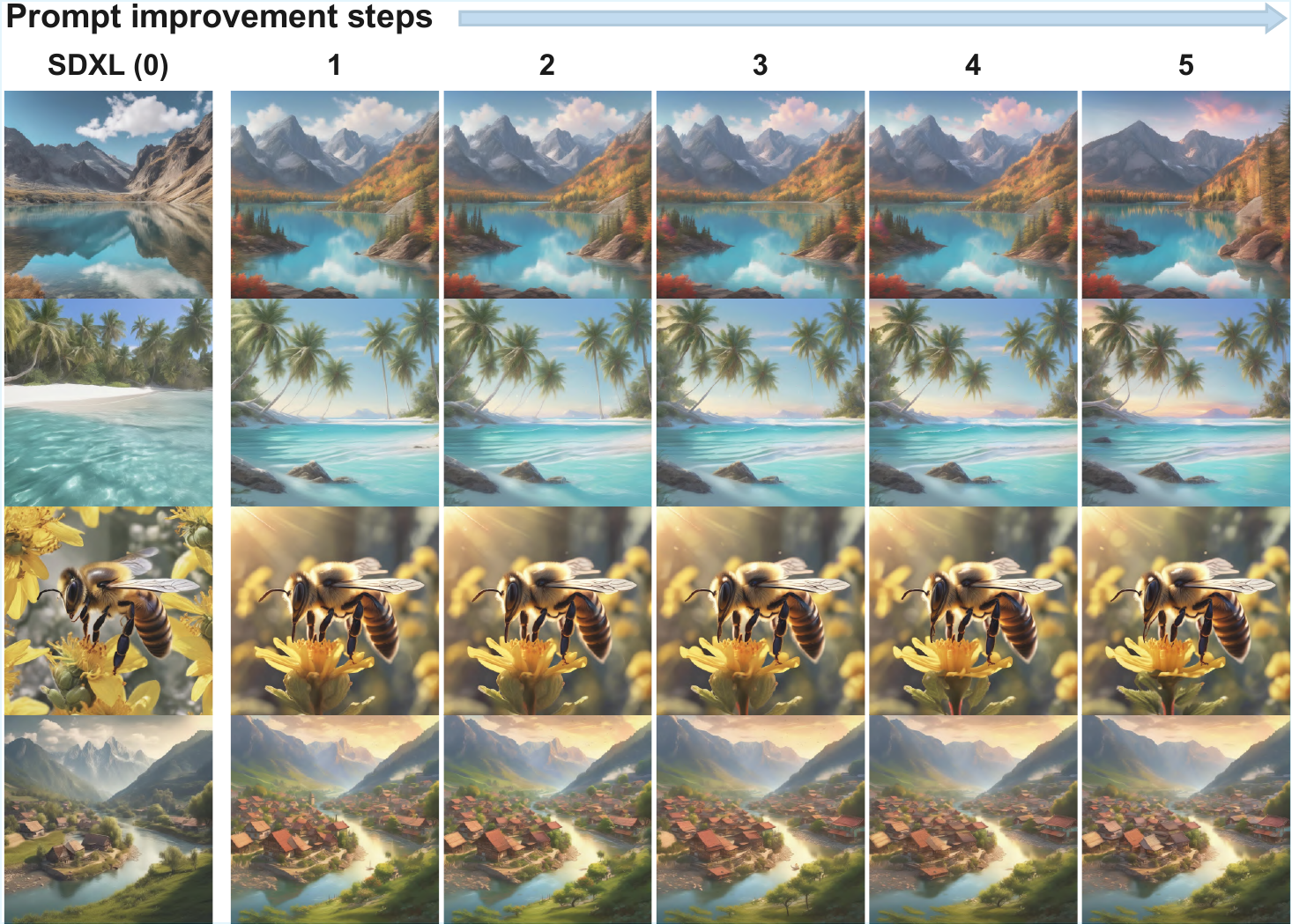}
\includegraphics[width=0.238\textwidth]{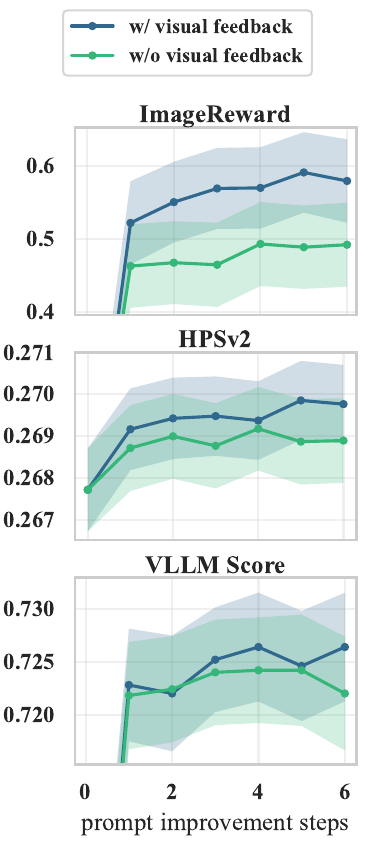}
\end{center}
\vspace{-0.5em}
\caption{Ablation study demonstrating that incorporating visual feedback and increasing the number of refinement steps consistently enhances reward alignment. (Left: SDXL, Right: SD1.5; reward: ImageReward)}
\label{fig:graph-abl}
\end{figure*}

We also investigated the impact of the number of prompt refinement steps (Fig.~\ref{fig:graph-abl}). Increasing the number of refinement steps improved not only the reward metric but also other evaluation metrics. Importantly, increasing the number of refinement steps does not increase the number of diffusion sampling steps. When trained without visual feedback, these improvements were much smaller or absent. These findings highlight that visual feedback and iterative prompt refinement are indispensable components of our equivalence MDP formulation. Together, they establish the closed-loop structure that mirrors direct RL on diffusion models, and the ablation results confirm that this formulation is not only structurally well-founded but also empirically effective.

For further analyses, including timestep-wise prompt evolution analysis and additional qualitative results, please refer to Appendix~\ref{supp-results} and \ref{supp-prompt-evolve}.

\begin{table}
\caption{Inference time analysis showing minimal runtime overhead of the proposed method (A100×1, batch size = 8).}
\label{tab:inference-time}
\centering
\resizebox{0.85\columnwidth}{!}
{
\small
\begin{tabular}{lccc}
\toprule
\textbf{Model} & \textbf{Improvement Steps}  & \multicolumn{2}{c}{\textbf{Inference Time}}  \\
& & s/img & relative \\
\toprule
 SDXL & 0 & 15.00 & 1.0 \\
 \midrule
 \multirow{5}{*}{\textbf{+ PromptLoop}} & 1 & 15.73 & 1.05 \\
 & 2 & 16.29 & 1.09 \\
 & 3 & 16.95 & 1.13 \\
 & 4 & 17.55 & 1.17 \\
 & 5 & 18.43 & 1.23 \\
\bottomrule
\end{tabular}
}
\end{table}

\noindent \textbf{Runtime Overhead.} To evaluate the runtime overhead introduced by the integrated MLLM policy model, we measure the inference time as a function of the number of prompt improvement steps. Tab.~\ref{tab:inference-time} shows that PromptLoop increases the total inference time only marginally, from 15s to 18s in our default configuration with five prompt refinement steps. This result indicates that our method keeps the inference-time overhead at a manageable level, demonstrating the user-level practicality of our framework.

Importantly, the MLLM policy model remains fixed in size and requires no retraining when applied to new diffusion backbones. As diffusion models continue to scale, the relative inference cost introduced by our policy model naturally decreases. This plug-and-play design thus offers broad architectural compatibility and long-term scalability, making the framework increasingly efficient and readily applicable to next-generation diffusion models without additional training cost.

\section{Conclusion}

In this work, we introduced PromptLoop, a plug-and-play framework for reward alignment of diffusion models via step-wise prompt refinement with latent feedback. By leveraging a multimodal policy model trained with reinforcement learning, our method attains structural equivalence to parameter-level fine-tuning while retaining the flexibility, generality, and modularity of prompt-based alignment. Experiments demonstrate that PromptLoop achieves effective reward optimization, generalizes seamlessly to unseen diffusion backbones, composes orthogonally with existing alignment techniques, and mitigates over-optimization and reward hacking. These results position PromptLoop not only as a structurally sound but also as a practically robust complement to weight-level tuning. Overall, PromptLoop provides a simple yet effective path toward more reliable and adaptable generative models, while its plug-and-play nature facilitates integration into user-facing applications, underscoring strong potential for real-world deployment.

\section*{Acknowledgment}
This work was supported by Institute of Information \& communications Technology Planning \& Evaluation (IITP) grant funded by the Korea government (MSIT) (No. RS-2022-II220984, Development of Artificial Intelligence Technology for Personalized Plug-and-Play Explanation and Verification of Explanation) and by the National Research Foundation of Korea under Grant RS-2024-00336454.

{
    \small
    \bibliographystyle{ieeenat_fullname}
    \bibliography{main}

@article{ho2020denoising,
  title={Denoising diffusion probabilistic models},
  author={Ho, Jonathan and Jain, Ajay and Abbeel, Pieter},
  journal={Advances in neural information processing systems},
  volume={33},
  pages={6840--6851},
  year={2020}
}

@article{song2020score,
  title={Score-based generative modeling through stochastic differential equations},
  author={Song, Yang and Sohl-Dickstein, Jascha and Kingma, Diederik P and Kumar, Abhishek and Ermon, Stefano and Poole, Ben},
  journal={arXiv preprint arXiv:2011.13456},
  year={2020}
}

@inproceedings{rombach2022high,
  title={High-resolution image synthesis with latent diffusion models},
  author={Rombach, Robin and Blattmann, Andreas and Lorenz, Dominik and Esser, Patrick and Ommer, Bj{\"o}rn},
  booktitle={Proceedings of the IEEE/CVF conference on computer vision and pattern recognition},
  pages={10684--10695},
  year={2022}
}

@article{schulman2017proximal,
  title={Proximal policy optimization algorithms},
  author={Schulman, John and Wolski, Filip and Dhariwal, Prafulla and Radford, Alec and Klimov, Oleg},
  journal={arXiv preprint arXiv:1707.06347},
  year={2017}
}

@article{rafailov2023direct,
  title={Direct preference optimization: Your language model is secretly a reward model},
  author={Rafailov, Rafael and Sharma, Archit and Mitchell, Eric and Manning, Christopher D and Ermon, Stefano and Finn, Chelsea},
  journal={Advances in neural information processing systems},
  volume={36},
  pages={53728--53741},
  year={2023}
}

@inproceedings{black2024training,
  title     = {Training Diffusion Models with Reinforcement Learning},
  author    = {Black, Kevin and Janner, Michael and Du, Yilun and Kostrikov, Ilya and Levine, Sergey},
  booktitle = {The Twelfth International Conference on Learning Representations},
  year      = {2024}
}

@inproceedings{wallace2024diffusion,
  title={Diffusion model alignment using direct preference optimization},
  author={Wallace, Bram and Dang, Meihua and Rafailov, Rafael and Zhou, Linqi and Lou, Aaron and Purushwalkam, Senthil and Ermon, Stefano and Xiong, Caiming and Joty, Shafiq and Naik, Nikhil},
  booktitle={Proceedings of the IEEE/CVF Conference on Computer Vision and Pattern Recognition},
  pages={8228--8238},
  year={2024}
}

@book{sutton1998reinforcement,
  title={Reinforcement learning: An introduction},
  author={Sutton, Richard S and Barto, Andrew G and others},
  volume={1},
  number={1},
  year={1998},
  publisher={MIT press Cambridge}
}

@inproceedings{kim2025test,
  title     = {Test-time Alignment of Diffusion Models without Reward Over-optimization},
  author    = {Kim, Sunwoo and Kim, Minkyu and Park, Dongmin},
  booktitle = {The Thirteenth International Conference on Learning Representations},
  year      = {2025}
}

@article{wu2025reprompt,
  title={RePrompt: Reasoning-Augmented Reprompting for Text-to-Image Generation via Reinforcement Learning},
  author={Wu, Mingrui and Wang, Lu and Zhao, Pu and Yang, Fangkai and Zhang, Jianjin and Liu, Jianfeng and Zhan, Yuefeng and Han, Weihao and Sun, Hao and Ji, Jiayi and others},
  journal={arXiv preprint arXiv:2505.17540},
  year={2025}
}

@article{brown2020language,
  title={Language models are few-shot learners},
  author={Brown, Tom and Mann, Benjamin and Ryder, Nick and Subbiah, Melanie and Kaplan, Jared D and Dhariwal, Prafulla and Neelakantan, Arvind and Shyam, Pranav and Sastry, Girish and Askell, Amanda and others},
  journal={Advances in neural information processing systems},
  volume={33},
  pages={1877--1901},
  year={2020}
}

@article{grattafiori2024llama,
  title={The llama 3 herd of models},
  author={Grattafiori, Aaron and Dubey, Abhimanyu and Jauhri, Abhinav and Pandey, Abhinav and Kadian, Abhishek and Al-Dahle, Ahmad and Letman, Aiesha and Mathur, Akhil and Schelten, Alan and Vaughan, Alex and others},
  journal={arXiv preprint arXiv:2407.21783},
  year={2024}
}

@article{guo2025deepseek,
  title={Deepseek-r1: Incentivizing reasoning capability in llms via reinforcement learning},
  author={Guo, Daya and Yang, Dejian and Zhang, Haowei and Song, Junxiao and Zhang, Ruoyu and Xu, Runxin and Zhu, Qihao and Ma, Shirong and Wang, Peiyi and Bi, Xiao and others},
  journal={arXiv preprint arXiv:2501.12948},
  year={2025}
}

@article{wang2024qwen2,
  title={Qwen2-vl: Enhancing vision-language model's perception of the world at any resolution},
  author={Wang, Peng and Bai, Shuai and Tan, Sinan and Wang, Shijie and Fan, Zhihao and Bai, Jinze and Chen, Keqin and Liu, Xuejing and Wang, Jialin and Ge, Wenbin and others},
  journal={arXiv preprint arXiv:2409.12191},
  year={2024}
}

@article{liu2023visual,
  title={Visual instruction tuning},
  author={Liu, Haotian and Li, Chunyuan and Wu, Qingyang and Lee, Yong Jae},
  journal={Advances in neural information processing systems},
  volume={36},
  pages={34892--34916},
  year={2023}
}

@article{wang2025internvl3,
  title={InternVL3. 5: Advancing Open-Source Multimodal Models in Versatility, Reasoning, and Efficiency},
  author={Wang, Weiyun and Gao, Zhangwei and Gu, Lixin and Pu, Hengjun and Cui, Long and Wei, Xingguang and Liu, Zhaoyang and Jing, Linglin and Ye, Shenglong and Shao, Jie and others},
  journal={arXiv preprint arXiv:2508.18265},
  year={2025}
}

@inproceedings{lester-etal-2021-power,
    title = "The Power of Scale for Parameter-Efficient Prompt Tuning",
    author = "Lester, Brian  and
      Al-Rfou, Rami  and
      Constant, Noah",
    editor = "Moens, Marie-Francine  and
      Huang, Xuanjing  and
      Specia, Lucia  and
      Yih, Scott Wen-tau",
    booktitle = "Proceedings of the 2021 Conference on Empirical Methods in Natural Language Processing",
    month = nov,
    year = "2021",
    address = "Online and Punta Cana, Dominican Republic",
    publisher = "Association for Computational Linguistics",
    url = "https://aclanthology.org/2021.emnlp-main.243/",
    doi = "10.18653/v1/2021.emnlp-main.243",
    pages = "3045--3059"
}

@article{genewein2025understanding,
  title={Understanding prompt tuning and in-context learning via meta-learning},
  author={Genewein, Tim and Li, Kevin Wenliang and Grau-Moya, Jordi and Ruoss, Anian and Orseau, Laurent and Hutter, Marcus},
  journal={arXiv preprint arXiv:2505.17010},
  year={2025}
}

@inproceedings{xie2022explanation,
  title     = {An Explanation of In-Context Learning as Implicit Bayesian Inference},
  author    = {Xie, Sang Michael and Raghunathan, Aditi and Liang, Percy and Ma, Tengyu},
  booktitle = {The Tenth International Conference on Learning Representations},
  year      = {2022}
}

@article{kim2025reward,
  title={Reward-Agnostic Prompt Optimization for Text-to-Image Diffusion Models},
  author={Kim, Semin and Cha, Yeonwoo and Yoo, Jaehoon and Hong, Seunghoon},
  journal={arXiv preprint arXiv:2506.16853},
  year={2025}
}

@article{hao2023optimizing,
  title={Optimizing prompts for text-to-image generation},
  author={Hao, Yaru and Chi, Zewen and Dong, Li and Wei, Furu},
  journal={Advances in Neural Information Processing Systems},
  volume={36},
  pages={66923--66939},
  year={2023}
}

@article{manas2024improving,
  title={Improving text-to-image consistency via automatic prompt optimization},
  author={Ma{\~n}as, Oscar and Astolfi, Pietro and Hall, Melissa and Ross, Candace and Urbanek, Jack and Williams, Adina and Agrawal, Aishwarya and Romero-Soriano, Adriana and Drozdzal, Michal},
  journal={arXiv preprint arXiv:2403.17804},
  year={2024}
}

@article{khan2025test, title={Test-time Prompt Refinement for Text-to-Image Models}, author={Khan, Mohammad Abdul Hafeez and Jain, Yash and Bhattacharyya, Siddhartha and Vineet, Vibhav}, journal={arXiv preprint arXiv:2507.22076}, year={2025} }

@article{vincent2011connection,
  title={A connection between score matching and denoising autoencoders},
  author={Vincent, Pascal},
  journal={Neural computation},
  volume={23},
  number={7},
  pages={1661--1674},
  year={2011},
  publisher={MIT Press}
}

@article{song2019generative,
  title={Generative modeling by estimating gradients of the data distribution},
  author={Song, Yang and Ermon, Stefano},
  journal={Advances in neural information processing systems},
  volume={32},
  year={2019}
}

@inproceedings{sohl2015deep,
  title={Deep unsupervised learning using nonequilibrium thermodynamics},
  author={Sohl-Dickstein, Jascha and Weiss, Eric and Maheswaranathan, Niru and Ganguli, Surya},
  booktitle={International conference on machine learning},
  pages={2256--2265},
  year={2015},
  organization={pmlr}
}

@article{chung2022diffusion,
  title={Diffusion posterior sampling for general noisy inverse problems},
  author={Chung, Hyungjin and Kim, Jeongsol and Mccann, Michael T and Klasky, Marc L and Ye, Jong Chul},
  journal={arXiv preprint arXiv:2209.14687},
  year={2022}
}

@inproceedings{yu2023freedom,
  title={Freedom: Training-free energy-guided conditional diffusion model},
  author={Yu, Jiwen and Wang, Yinhuai and Zhao, Chen and Ghanem, Bernard and Zhang, Jian},
  booktitle={Proceedings of the IEEE/CVF International Conference on Computer Vision},
  pages={23174--23184},
  year={2023}
}

@article{williams1992simple,
  title={Simple statistical gradient-following algorithms for connectionist reinforcement learning},
  author={Williams, Ronald J},
  journal={Machine learning},
  volume={8},
  number={3},
  pages={229--256},
  year={1992},
  publisher={Springer}
}

@article{sutton1999policy,
  title={Policy gradient methods for reinforcement learning with function approximation},
  author={Sutton, Richard S and McAllester, David and Singh, Satinder and Mansour, Yishay},
  journal={Advances in neural information processing systems},
  volume={12},
  year={1999}
}

@article{song2020denoising,
  title={Denoising diffusion implicit models},
  author={Song, Jiaming and Meng, Chenlin and Ermon, Stefano},
  journal={arXiv preprint arXiv:2010.02502},
  year={2020}
}

@inproceedings{liu2022pseudo,
  title     = {Pseudo Numerical Methods for Diffusion Models on Manifolds},
  author    = {Liu, Luping and Ren, Yi and Lin, Zhijie and Zhao, Zhou},
  booktitle = {The Tenth International Conference on Learning Representations},
  year      = {2022}
}

@article{karras2022elucidating,
  title={Elucidating the design space of diffusion-based generative models},
  author={Karras, Tero and Aittala, Miika and Aila, Timo and Laine, Samuli},
  journal={Advances in neural information processing systems},
  volume={35},
  pages={26565--26577},
  year={2022}
}

@article{lu2022dpm,
  title={Dpm-solver: A fast ode solver for diffusion probabilistic model sampling in around 10 steps},
  author={Lu, Cheng and Zhou, Yuhao and Bao, Fan and Chen, Jianfei and Li, Chongxuan and Zhu, Jun},
  journal={Advances in neural information processing systems},
  volume={35},
  pages={5775--5787},
  year={2022}
}

@misc{schuhmann_improved_aesthetic_predictor_2025,
  title        = {{CLIP+MLP Aesthetic Score Predictor}},
  author       = {Schuhmann, Christoph},
  howpublished = {GitHub repository},
  year         = {2025},
  note         = {\url{https://github.com/christophschuhmann/improved-aesthetic-predictor}},
}

@inproceedings{radford2021learning,
  title={Learning transferable visual models from natural language supervision},
  author={Radford, Alec and Kim, Jong Wook and Hallacy, Chris and Ramesh, Aditya and Goh, Gabriel and Agarwal, Sandhini and Sastry, Girish and Askell, Amanda and Mishkin, Pamela and Clark, Jack and others},
  booktitle={International conference on machine learning},
  pages={8748--8763},
  year={2021},
  organization={PmLR}
}

@article{wu2023human,
  title={Human preference score v2: A solid benchmark for evaluating human preferences of text-to-image synthesis},
  author={Wu, Xiaoshi and Hao, Yiming and Sun, Keqiang and Chen, Yixiong and Zhu, Feng and Zhao, Rui and Li, Hongsheng},
  journal={arXiv preprint arXiv:2306.09341},
  year={2023}
}

@article{xu2023imagereward,
  title={Imagereward: Learning and evaluating human preferences for text-to-image generation},
  author={Xu, Jiazheng and Liu, Xiao and Wu, Yuchen and Tong, Yuxuan and Li, Qinkai and Ding, Ming and Tang, Jie and Dong, Yuxiao},
  journal={Advances in Neural Information Processing Systems},
  volume={36},
  pages={15903--15935},
  year={2023}
}

@misc{laion_safety,
  title        = {LAION SAFETY: {CLIP}-based NSFW Detection},
  author       = {{LAION-AI}},
  year         = {2023},
  howpublished = {GitHub repository},
  note         = {Available at \url{https://github.com/LAION-AI/LAION-SAFETY}},
}

@article{bai2025qwen2,
  title={Qwen2. 5-vl technical report},
  author={Bai, Shuai and Chen, Keqin and Liu, Xuejing and Wang, Jialin and Ge, Wenbin and Song, Sibo and Dang, Kai and Wang, Peng and Wang, Shijie and Tang, Jun and others},
  journal={arXiv preprint arXiv:2502.13923},
  year={2025}
}

@article{podell2023sdxl,
  title={Sdxl: Improving latent diffusion models for high-resolution image synthesis},
  author={Podell, Dustin and English, Zion and Lacey, Kyle and Blattmann, Andreas and Dockhorn, Tim and M{\"u}ller, Jonas and Penna, Joe and Rombach, Robin},
  journal={arXiv preprint arXiv:2307.01952},
  year={2023}
}

@inproceedings{sauer2024adversarial,
  title={Adversarial diffusion distillation},
  author={Sauer, Axel and Lorenz, Dominik and Blattmann, Andreas and Rombach, Robin},
  booktitle={European Conference on Computer Vision},
  pages={87--103},
  year={2024},
  organization={Springer}
}

@article{kirstain2023pick,
  title={Pick-a-pic: An open dataset of user preferences for text-to-image generation},
  author={Kirstain, Yuval and Polyak, Adam and Singer, Uriel and Matiana, Shahbuland and Penna, Joe and Levy, Omer},
  journal={Advances in neural information processing systems},
  volume={36},
  pages={36652--36663},
  year={2023}
}

@misc{openai_gpt5mini_doc_2025,
  title        = {OpenAI Platform: gpt-5-mini},
  author       = {{OpenAI}},
  year         = {2025},
  howpublished = {\url{https://platform.openai.com/docs/models/gpt-5-mini}},
  note         = {Accessed on 2025-09-14},
}

@misc{openai_gpt5_doc_2025,
  title        = {OpenAI Platform: gpt-5},
  author       = {{OpenAI}},
  year         = {2025},
  howpublished = {\url{https://platform.openai.com/docs/models/gpt-5}},
  note         = {Accessed on 2025-11-13},
}

@inproceedings{zhou2025golden,
      title={Golden Noise for Diffusion Models: A Learning Framework}, 
      author={Zikai Zhou and Shitong Shao and Lichen Bai and Shufei Zhang and Zhiqiang Xu and Bo Han and Zeke Xie},
      booktitle={International Conference on Computer Vision},
      year={2025},
}

@article{ghosh2023geneval,
  title={Geneval: An object-focused framework for evaluating text-to-image alignment},
  author={Ghosh, Dhruba and Hajishirzi, Hannaneh and Schmidt, Ludwig},
  journal={Advances in Neural Information Processing Systems},
  volume={36},
  pages={52132--52152},
  year={2023}
}

@inproceedings{hu2022lora,
  title     = {Lora: Low-rank adaptation of large language models},
  author    = {Hu, Edward J and Shen, Yelong and Wallis, Phillip and Allen-Zhu, Zeyuan and Li, Yuanzhi and Wang, Shean and Wang, Lu and Chen, Weizhu and others},
  booktitle = {The Tenth International Conference on Learning Representations},
  year      = {2022}
}

@article{xue2025dancegrpo,
  title={DanceGRPO: Unleashing GRPO on Visual Generation},
  author={Xue, Zeyue and Wu, Jie and Gao, Yu and Kong, Fangyuan and Zhu, Lingting and Chen, Mengzhao and Liu, Zhiheng and Liu, Wei and Guo, Qiushan and Huang, Weilin and others},
  journal={arXiv preprint arXiv:2505.07818},
  year={2025}
}

@article{hu2024ella,
  title={Ella: Equip diffusion models with llm for enhanced semantic alignment},
  author={Hu, Xiwei and Wang, Rui and Fang, Yixiao and Fu, Bin and Cheng, Pei and Yu, Gang},
  journal={arXiv preprint arXiv:2403.05135},
  year={2024}
}

@article{domingo2024adjoint,
  title={Adjoint matching: Fine-tuning flow and diffusion generative models with memoryless stochastic optimal control},
  author={Domingo-Enrich, Carles and Drozdzal, Michal and Karrer, Brian and Chen, Ricky TQ},
  journal={arXiv preprint arXiv:2409.08861},
  year={2024}
}

@article{wang2025promptenhancer,
  title={PromptEnhancer: A Simple Approach to Enhance Text-to-Image Models via Chain-of-Thought Prompt Rewriting},
  author={Wang, Linqing and Xing, Ximing and Cheng, Yiji and Zhao, Zhiyuan and Tao, Jiale and Wang, Qixun and Li, Ruihuang and Li, Xin and Wu, Mingrui and Deng, Xinchi and others},
  journal={arXiv preprint arXiv:2509.04545},
  year={2025}
}

@article{li2025visual,
  title={Visual-CoG: Stage-Aware Reinforcement Learning with Chain of Guidance for Text-to-Image Generation},
  author={Li, Yaqi and Chen, Peng and Han, Mingyang and Pi, Bu and Shi, Haoxiang and Zhao, Runzhou and Yao, Yang and Zhang, Xuan and Song, Jun},
  journal={arXiv preprint arXiv:2508.18032},
  year={2025}
}

@article{huang2025interleaving,
  title={Interleaving Reasoning for Better Text-to-Image Generation},
  author={Huang, Wenxuan and Chen, Shuang and Xie, Zheyong and Cao, Shaosheng and Tang, Shixiang and Shen, Yufan and Yin, Qingyu and Hu, Wenbo and Wang, Xiaoman and Tang, Yuntian and others},
  journal={arXiv preprint arXiv:2509.06945},
  year={2025}
}

@article{kynkaanniemi2024applying,
  title={Applying guidance in a limited interval improves sample and distribution quality in diffusion models},
  author={Kynk{\"a}{\"a}nniemi, Tuomas and Aittala, Miika and Karras, Tero and Laine, Samuli and Aila, Timo and Lehtinen, Jaakko},
  journal={Advances in Neural Information Processing Systems},
  volume={37},
  pages={122458--122483},
  year={2024}
}

@article{wang2024analysis,
  title={Analysis of classifier-free guidance weight schedulers},
  author={Wang, Xi and Dufour, Nicolas and Andreou, Nefeli and Cani, Marie-Paule and Abrevaya, Victoria Fern{\'a}ndez and Picard, David and Kalogeiton, Vicky},
  journal={arXiv preprint arXiv:2404.13040},
  year={2024}
}

@article{papalampidi2025dynamic,
  title={Dynamic Classifier-Free Diffusion Guidance via Online Feedback},
  author={Papalampidi, Pinelopi and Wiles, Olivia and Ktena, Ira and Shtedritski, Aleksandar and Bugliarello, Emanuele and Kajic, Ivana and Albuquerque, Isabela and Nematzadeh, Aida},
  journal={arXiv preprint arXiv:2509.16131},
  year={2025}
}

@inproceedings{kim2025free2guide,
  title={Free2Guide: Training-Free Text-to-Video Alignment using Image LVLM},
  author={Kim, Jaemin and Kim, Bryan Sangwoo and Ye, Jong Chul},
  booktitle={Proceedings of the IEEE/CVF International Conference on Computer Vision},
  pages={17920--17929},
  year={2025}
}

@inproceedings{yang2024idea2img,
  title={Idea2img: Iterative self-refinement with gpt-4v for automatic image design and generation},
  author={Yang, Zhengyuan and Wang, Jianfeng and Li, Linjie and Lin, Kevin and Lin, Chung-Ching and Liu, Zicheng and Wang, Lijuan},
  booktitle={European Conference on Computer Vision},
  pages={167--184},
  year={2024},
  organization={Springer}
}

@article{singhal2025general,
  title={A general framework for inference-time scaling and steering of diffusion models},
  author={Singhal, Raghav and Horvitz, Zachary and Teehan, Ryan and Ren, Mengye and Yu, Zhou and McKeown, Kathleen and Ranganath, Rajesh},
  journal={arXiv preprint arXiv:2501.06848},
  year={2025}
}

@InProceedings{chung2022improving,
  title={Improving Diffusion Models for Inverse Problems using Manifold Constraints},
  author={Chung, Hyungjin and Sim, Byeongsu and Ye, Jong Chul},
  journal={Advances in Neural Information Processing Systems},
  year={2022}
}

@misc{flux1dev2024,
  author       = {Black Forest Labs},
  title        = {FLUX.1-dev},
  year         = {2024},
  howpublished = {\url{https://huggingface.co/black-forest-labs/FLUX.1-dev}},
  note         = {Text-to-image diffusion model}
}

@inproceedings{esser2024scaling,
  title={Scaling rectified flow transformers for high-resolution image synthesis},
  author={Esser, Patrick and Kulal, Sumith and Blattmann, Andreas and Entezari, Rahim and M{\"u}ller, Jonas and Saini, Harry and Levi, Yam and Lorenz, Dominik and Sauer, Axel and Boesel, Frederic and others},
  booktitle={Forty-first international conference on machine learning},
  year={2024}
}

@inproceedings{bert-score,
  title={BERTScore: Evaluating Text Generation with BERT},
  author={Tianyi Zhang* and Varsha Kishore* and Felix Wu* and Kilian Q. Weinberger and Yoav Artzi},
  booktitle={International Conference on Learning Representations},
  year={2020},
  url={https://openreview.net/forum?id=SkeHuCVFDr}
}

@inproceedings{
he2021deberta,
title={DEBERTA: DECODING-ENHANCED BERT WITH DISENTANGLED ATTENTION},
author={Pengcheng He and Xiaodong Liu and Jianfeng Gao and Weizhu Chen},
booktitle={International Conference on Learning Representations},
year={2021},
url={https://openreview.net/forum?id=XPZIaotutsD}
}
}


\clearpage
\onecolumn
\setcounter{page}{1}
\appendix
\section*{\centering \LARGE \textbf{Appendix}}
\section*{\centering \textbf{Supplementary Materials}}

\section{Related Works} \label{supp-related-works}

\textbf{Aligning Diffusion Models.}  
Following the success of RLHF for LLMs, there has been growing interest in aligning diffusion models with human preferences or arbitrary reward functions. Methods such as DDPO~\citep{black2024training}, Diffusion-DPO~\citep{wallace2024diffusion}, and DanceGRPO~\citep{xue2025dancegrpo} treat the diffusion sampling process as a Markov decision process (MDP), and train the diffusion model using RL algorithms. In contrast to RL-based approaches that rely on black-box rewards, other methods directly exploit the gradient of the reward or objective function. For example, ReFL~\citep{xu2023imagereward} optimizes sampling trajectories via reward gradients, applying the reward to intermediate denoised estimates to avoid full backpropagation. ELLA~\citep{hu2024ella} introduces a timestep-aware connector module that maps encoded prompt embeddings before they are fed into the diffusion model. More recently, Adjoint Matching~\citep{domingo2024adjoint} casts reward fine-tuning as a stochastic optimal control (SOC) problem, optimizing with reward gradients.

\noindent \textbf{Prompt-based Improvements for Diffusion Models.}  
In text-to-image generation, prompts serve as a powerful control signal and have been widely leveraged as a means of alignment. Prior work such as OPT2I~\citep{manas2024improving}, Idea2Img~\cite{yang2024idea2img}, RATTPO~\citep{kim2025reward}, and TIR~\citep{khan2025test} explores LLM-based prompt refinement without fine-tuning, relying on feedback from evaluations of fully generated images to suggest improved prompts. To align LLM-based prompt refinement more closely with reward, Promptist~\citep{hao2023optimizing}, RePrompt~\citep{wu2025reprompt}, and PromptEnhancer~\citep{wang2025promptenhancer} fine-tune LLMs with reinforcement learning, treating the diffusion model simply as a black-box reward model in a feedforward manner. RL-based alignment has also been extended beyond diffusion models to autoregressive (AR) multimodal models, where methods such as Visual-CoG~\citep{li2025visual} and IRGL~\citep{huang2025interleaving} adopt CoT-style approaches that iteratively generate prompts and images through self-feedback to achieve reward alignment.

\section{Detailed Algorithm}

We summarize the procedure of PromptLoop in two parts. Algorithm~\ref{alg:training} presents the training process, while Algorithm~\ref{alg:sampling} details the sampling procedure.

\begin{algorithm}
\caption{Training PromptLoop}
\label{alg:training}
\DontPrintSemicolon
\SetAlgoLined
\KwIn{Policy $\pi_\theta$, diffusion denoiser $\hat{\boldsymbol{\epsilon}}_\phi$, sampler $f$, prompts $p_{\text{data}}$, reward $R$, \# refinement steps $N_R$, GRPO group size $G$, total steps $T$}
\KwOut{Reward-aligned plug-and-play policy $\pi_\theta$}

\Repeat{optimization complete}{
  Sample $q \sim p_{\text{data}}$\;
  Sample $\mathcal{R} \sim \mathrm{Unif}\!\left(\{\,R \subseteq \{1,\dots,T\}\,:\,|R|=N_R\,\}\right)$\;

  \For{$g \in \{1,\dots,G\}$}{
    $\boldsymbol{c} \gets q$ \tcp*{init text prompt}
    $\tau^g \gets [\,]$ \tcp*{trajectory: (state, action) pairs}
    Sample $\boldsymbol{x}_T \sim \mathcal{N}(0,I)$\;

    \For{$t = T, T-1, \dots, 1$}{
      \If{$t \in \mathcal{R}$}{
        $s_t \gets (\hat{\boldsymbol{x}}_t,\boldsymbol{c},q,t)$\;
        Sample $\boldsymbol{c} \sim \pi_\theta(\cdot \mid s_t)$ \tcp*{prompt refinement}
        $\tau^g.\text{append}(s_t)$; \quad $\tau^g.\text{append}(\boldsymbol{c})$\;
      }

      \tcp{perform one sampler step}
      Sample $\boldsymbol{z}_t \sim \mathcal{N}(0,I)$\;
      $\boldsymbol{x}_{t-1} \gets f(\boldsymbol{x}_t, \bm{z}_t, \bm{c}, t)$\;

      $\hat{\boldsymbol{x}}_{t-1} \gets \dfrac{1}{\sqrt{\bar{\alpha}_t}}\!\left(\boldsymbol{x}_t - \sqrt{1-\bar{\alpha}_t}\, \hat{\boldsymbol{\epsilon}}_\phi(\boldsymbol{x}_t,t,\boldsymbol{c})\right)$\;
    }

    $r^g \gets R(\boldsymbol{x}_0, q)$ \tcp*{reward calculation}
  }

  Update $\pi_\theta$ with GRPO using $\{(\tau^g, r^g)\}_{g=1}^G$\;
}
\end{algorithm}

\begin{algorithm}
\caption{Sampling with PromptLoop}
\label{alg:sampling}
\DontPrintSemicolon
\SetAlgoLined

\KwIn{Policy $\pi_\theta$, diffusion denoiser $\hat{\boldsymbol{\epsilon}}_\phi$, sampler $f$, input prompt $q$, refinement steps $\mathcal{R}\subseteq\{1,\dots,T\}$}
\KwOut{Reward-aligned sample $\boldsymbol{x}_0$}

Sample $\boldsymbol{x}_T \sim \mathcal{N}(0, I)$\;
$\boldsymbol{c} \gets q$\;

\For{$t = T, T-1, \dots, 1$}{
  \If{$t \in \mathcal{R}$}{
    $s_t \gets (\hat{\boldsymbol{x}}_t, \boldsymbol{c}, q, t)$\;
    Sample $\boldsymbol{c} \sim \pi_\theta(\cdot \mid s_t)$ \tcp*{prompt refinement}
  }

  Sample $\boldsymbol{z}_t \sim \mathcal{N}(0,I)$\;
  $\boldsymbol{x}_{t-1} \gets f(\boldsymbol{x}_t, \bm{z}_t, \bm{c}, t)$\;

  $\hat{\boldsymbol{x}}_{t-1} \gets \dfrac{1}{\sqrt{\bar{\alpha}_t}}\!\left(\boldsymbol{x}_t - \sqrt{1-\bar{\alpha}_t}\, \hat{\boldsymbol{\epsilon}}_\phi(\boldsymbol{x}_t,t,\boldsymbol{c})\right)$\;
}
\end{algorithm}

\section{Implementation Details}

\subsection{Framework and Training} 

We use Qwen2.5-VL-3B-Instruct~\citep{bai2025qwen2} as the policy model, and Stable Diffusion 1.5~\citep{rombach2022high} (SD1.5), XL~\citep{podell2023sdxl} (SDXL), and XL-Turbo~\citep{sauer2024adversarial} (SDXL-turbo) as the text-to-image diffusion backbones, with the specific model chosen according to the task setting. Generation resolution, classifier-free guidance (CFG) scale, inference steps, and sampler were set to each model’s default configuration, except that we used the DDIM sampler~\citep{song2020denoising} for SD1.5 and 5 sampling steps for SDXL-turbo.

For GRPO training, we build on the TRL library\footnote{\url{https://github.com/huggingface/trl}} and implement our framework on top of it. Training is performed with the GRPO algorithm using a learning rate of $5\times10^{-6}$, batch size $8$, group size $8$, and $\beta$ (the KL-regularization coefficient) set to $0.005$ for single-reward training and $0$ for composite-reward training, without PPO clipping (num-iterations $=1$). We further apply parameter-efficient fine-tuning (LoRA)~\citep{hu2022lora} using the PEFT library\footnote{\url{https://github.com/huggingface/peft}}, with rank $r=16$, scaling factor $\alpha=64$, dropout $0.05$, and updates applied to all linear projection layers in the transformer blocks. All experiments are conducted in \texttt{bf16} precision on four NVIDIA A100 80GB GPUs, and each training run takes approximately three days to complete.  

To optimize our framework, we use $2$ training-prompt improvement steps and $5$ sampling-prompt improvement steps. Visual feedback is resized to $256 \times 256$ from the original denoised estimates obtained during the sampling process and provided to the policy model. During sampling, we insert the built-in token \texttt{<|image\_pad|>} as a placeholder to replace the visual feedback.

\subsection{Prompting Policy Models}

The policy models used for prompt refinement are guided by the instruction shown in Fig.~\ref{fig:vllm-policy-prompt},~\ref{fig:vllm-policy-prompt-composite}. As described earlier, the policy model is conditioned on the raw user input, the previously applied improved prompt, and the current timestep. In addition, we provide auxiliary information such as the total number of timesteps and the name of the target reward function. The model is then required to output an improved prompt that is suitable for the current denoising step. For the reward specification, we only provide the name of the reward (\textit{e.g.}, \texttt{ImageReward}, \texttt{HPSv2}), without detailed definitions. This design leaves open the possibility of using the reward identifier as a mechanism for multi-reward alignment in future work. For composite rewards, the increased complexity results in longer prompts, which can hinder the diffusion model’s responsiveness. To address this, we employ a dedicated prompt design that explicitly accounts for this issue.

\begin{figure}[!ht]
  \centering
  \small
  \begin{tcolorbox}[
    colback=gray!5!white,
    colframe=black,
    title=\textbf{Policy Model Prompt (Single Reward)}
  ]

    \textbf{User Prompt:} 

    You are helping to refine a prompt for an image generation diffusion model. At each timestep, you are given the input prompt, lastly improved prompt with timestep, current timestep, total timesteps, a target reward function, and the partially generated image at the current diffusion timestep. Your task is to suggest an improved prompt that better aligns with the goal. Do not attempt to correct blurriness, as the partially generated image is expected to be unclear during diffusion.
    \\\\
    Respond \textit{only} with a valid JSON object in the following format without any other text:
\begin{verbatim}
{
  "improved_prompt": "<your improved prompt string>"
}

\end{verbatim}
Input:
\begin{verbatim}
{
  "input_prompt": {input_prompt},
  "last_prompt": {applied_prompt},
  "target_reward": {target_reward},
  "current_timestep": {current_timestep},
  "total_timesteps": {total_timesteps},
}
\end{verbatim}
  \end{tcolorbox}
  \caption{Prompt provided to the policy model for refinement. The instruction specifies the available context (user input, last improved prompt, timestep information, and reward name), and the model must output an improved prompt in JSON format.}
  \label{fig:vllm-policy-prompt}
\end{figure}

\begin{figure}[!ht]
  \centering
  \small
  \begin{tcolorbox}[
    colback=gray!5!white,
    colframe=black,
    title=\textbf{Policy Model Prompt (Composite Reward)}
  ]

    \textbf{User Prompt:} 

    You are helping to refine a prompt for an image generation diffusion model. \\\\
    
    [IMPORTANT] However, you must make \textit{minimal changes} to the original user's input and \textit{keep the prompt as simple as possible}. I \textit{strongly} recommend \textit{not modifying} the input prompt if possible. [IMPORTANT] \\\\
    
    Respond \textit{only} with a valid JSON object in the following format without any other text:

\begin{verbatim}
{
  "improved_prompt": "<your improved prompt string>"
}

\end{verbatim}
Input:
\begin{verbatim}
{
  "input_prompt": {input_prompt},
  "last_prompt": {applied_prompt},
  "target_reward": {target_reward},
  "current_timestep": {current_timestep},
  "total_timesteps": {total_timesteps},
}
\end{verbatim}
  \end{tcolorbox}
  \caption{Prompt provided to the policy model for refinement. The instruction specifies the available context (user input, last improved prompt, timestep information, and reward name), and the model must output an improved prompt in JSON format.}
  \label{fig:vllm-policy-prompt-composite}
\end{figure}

\subsection{Reward Models}

In the single-reward setting, we used ImageReward~\citep{xu2023imagereward}, incompressibility~\citep{black2024training}, compressibility~\citep{black2024training}, and aesthetic score models~\citep{schuhmann_improved_aesthetic_predictor_2025} without any modification from their official implementations and checkpoints. For the composite reward in the RePrompt-style setting, we adopted the same components—visual reasoning, length, and structure rewards. The visual reasoning reward consists of ImageReward and an MLLM-based reward, weighted equally, where the latter is implemented with \texttt{gpt-5-mini-2025-08-07}~\citep{openai_gpt5mini_doc_2025}. The evaluation prompt for the MLLM reward is shown in Fig.~\ref{fig:vllm-reward-prompt}. This design complements ImageReward by preventing reward hacking related to weak text alignment and aesthetic biases. The length reward follows the original formulation without change, while the structure reward is adapted to match our output format (JSON). Across all reward components, the scoring ranges and configurations remain unchanged.

\begin{figure}[!ht]
  \centering
  \small
  \begin{tcolorbox}[
    colback=gray!5!white,
    colframe=black,
    title=\textbf{MLLM Reward Model Prompt}
  ]

    \textbf{User Prompt:} 
You are an expert evaluator of text-to-image alignment. Your primary goal is to check whether the image faithfully matches the input prompt. Pay special attention to object identity, count, attributes (such as color, size, shape), and spatial relationships.\\
Penalize any elements that are not requested in the prompt — unnecessary decorations, background additions, or irrelevant visual noise. Missing or incorrect objects should also lower the score.\\
The best images are object-centric: focused on the entities and relationships specified in the prompt, while also being visually coherent and pleasant.\\\\
Please rate this image on a scale of 0-10 (10 being perfect) and explain your reasoning. Please put your score in $<$score$>$ score $<$/score$>$. Prompt: \{p\}
  \end{tcolorbox}
  \caption{Prompt template for the MLLM reward in the RePrompt-style composite setting, guiding fine-grained alignment checks and producing a structured score.}
  \label{fig:vllm-reward-prompt}
\end{figure}

\subsection{Evaluations}

\noindent \textbf{Baselines.} We use the official public PyTorch implementations of DDPO\footnote{\url{https://github.com/kvablack/ddpo-pytorch}} and ReFL\footnote{\url{https://github.com/zai-org/ImageReward}}, training them on the same dataset and reward model as PromptLoop. For ReFL on SD1.5, we perform full model fine-tuning, whereas for DDPO and ReFL on SDXL we adopt LoRA-based training. Reported performance values correspond to checkpoints where evaluation rewards match those of PromptLoop. Qwen2.5-VL-3B and GPT-5 (\texttt{gpt-5-2025-08-07}) are incorporated without GRPO training, relying solely on prompting (including visual feedback and multi-turn refinement), while maintaining the overall framework. RePrompt is implemented by removing visual feedback and multi-turn refinement from PromptLoop; reasoning is also omitted to ensure fair comparison under equivalent conditions. For Diffusion-DPO\footnote{\url{https://github.com/SalesforceAIResearch/DiffusionDPO}} and NPNet\footnote{\url{https://github.com/xie-lab-ml/Golden-Noise-for-Diffusion-Models}}, we directly used their officially released checkpoints and inference code without modification. For DanceGRPO,\footnote{\url{https://github.com/XueZeyue/DanceGRPO}} we reproduce its results using the official training code and dataset, training for 50 epochs with the HPSv2~\cite{wu2023human} reward model.

\noindent \textbf{Metrics.} For the single-reward setting, we evaluate models using ImageReward~\citep{xu2023imagereward}, HPSv2~\citep{wu2023human}, and an aesthetic scoring model~\citep{schuhmann_improved_aesthetic_predictor_2025}. These metrics assess prompt alignment, consistency with human preference, and robustness to over-optimization. We follow the standard evaluation protocols provided in the public implementations without any modifications.

In addition, we compute MLLM scores using a pretrained multimodal large language model, Qwen2.5-VL-3B-Instruct~\citep{wang2024qwen2}. 
The evaluation is performed locally with carefully designed prompts that balance human-preference alignment and aesthetic quality. 
Input images are resized to $512\times512$ before being fed into the model. 
The evaluator is instructed to provide a score between 0 and 10, with 10 indicating perfect quality. 
Scores are subsequently normalized to the range $[0,1]$ during post-processing. 
The full evaluation prompt is shown in Fig.~\ref{fig:vllm-metric-prompt}. 

For all these metrics, the evaluation prompts are drawn from the validation split of the Pick-a-Pic v2 dataset.

\begin{figure}[!ht]
  \centering
  \small
  \begin{tcolorbox}[
    colback=gray!5!white,
    colframe=black,
    title=\textbf{MLLM Score Metric Prompt}
  ]

    \textbf{User Prompt:} 

    You are an expert image evaluator. Your task is to judge an image based on two equally weighted aspects: \\ \\
    1. \textit{Faithfulness to Prompt}: Does the image accurately reflect the user’s input prompt in terms of objects, attributes, style, and composition? \\
    2. \textit{Aesthetic Quality}: Is the image visually appealing, well-composed, and artistically pleasant from a human perspective? \\ \\
    Please rate this image on a scale of 0-10 (10 being perfect) and explain your reasoning. Please put your score in $<
    $score$>$ score $<$/score$>$. Prompt: \{prompt\}
   
  \end{tcolorbox}
\caption{Evaluation prompt used for computing MLLM scores. The scoring model jointly considers prompt faithfulness and aesthetic quality, and outputs a rating from 0 to 10, which is subsequently normalized to the range [0, 1] in a post-processing step.}
  \label{fig:vllm-metric-prompt}
\end{figure}

In the composite-reward setting, we additionally evaluate on the GenEval benchmark~\citep{ghosh2023geneval}, which emphasizes object-centric aspects of text-to-image generation. We directly adopt the prompts and evaluation procedures provided by the GenEval benchmark without modification. When measuring ImageReward and HPSv2, we also use the prompts and the sample counts from GenEval.

\section{Prompt Evolution Analysis} \label{supp-prompt-evolve}

\subsection{Quantitative Analysis of Prompt Evolution}

Since our method controls the sampling dynamics of the diffusion model through textual prompts, the evolution trajectory over diffusion timesteps optimized via reinforcement learning remains interpretable, unlike \citet{hu2024ella}. To analyze this, we examine the outputs of a policy model trained on SDXL with ImageReward as a single reward signal. Tab.~\ref{tab:prompt-evolve} illustrates how the optimized prompts evolve as the diffusion timesteps progress.

Not every case follows the exact same trajectory, but a consistent overall pattern emerges across examples. At early timesteps, prompts typically emphasize meta-level descriptors highlighting quality, style, and realism (\textit{e.g.}, “photorealistic,” “vivid colors”), establishing a broad atmospheric framing. As inference advances to intermediate timesteps, these high-level descriptors give way to more concrete and fine-grained details, such as object properties, environmental elements, or specific lighting conditions, resulting in richer and more grounded descriptions. Toward later timesteps, we observe two dominant tendencies: in some cases, prompts continue to preserve the specificity around salient elements of the scene, while in others they collapse back into prototypical atmospheric cues (\textit{e.g.}, “warm glow,” “serene atmosphere”). This overall progression—from evaluative abstraction, to concrete specificity, and finally toward either preserved details or prototypical generalities—highlights how reinforcement-learned prompt evolution balances descriptive richness with compact, high-level guidance throughout the diffusion trajectory.

Interestingly, the RL-optimized prompt evolvement trajectory aligns with well-known scheduling strategies of classifier-free guidance (CFG). In diffusion models, it is established that the early steps focus on generating coarse global structures, while later steps refine finer details~\citep{yu2023freedom}. Consistent with this, prior studies have demonstrated that applying a strong CFG too early can be harmful, leading to a variety of scheduling strategies. Two dominant families of approaches exist: those that monotonically increase CFG strength throughout the sampling process and those that increase CFG up to intermediate timesteps before decreasing it again toward the final steps~\citep{wang2024analysis, kynkaanniemi2024applying, papalampidi2025dynamic}. Since stronger CFG effectively enforces sharper and more detailed conditioning, our results suggest that the RL-trained policy implicitly learns both types of dynamics at the textual level, adapting prompt specificity in ways that mirror optimal CFG schedules. This emergent behavior, despite not being explicitly instructed, is intriguing.

\begin{table}[t]
\caption{Comparative analysis of prompt evolvement at different timesteps. Early prompts emphasize broad atmospheric qualities, intermediate prompts expand into concrete details, and later prompts either preserve these specifics or revert to prototypical descriptors.}
\label{tab:prompt-evolve}
\centering
\resizebox{1.0\textwidth}{!}
{
\small
\begin{tabular}{lp{0.3\textwidth}p{0.3\textwidth}p{0.3\textwidth}}
\toprule
& \textbf{Initial ($t=981.0$)} & \textbf{Middle ($t=581.0$)} & \textbf{Final ($t=181.0$)} \\
\midrule
\textbf{Corgi Dog} & ...corgi wearing a hat and sunglasses, sitting on a beach chair, with a \textbf{picturesque beach and ocean in the background}. & ...corgi puppy wearing a multicolored bucket hat and sunglasses, sitting on a \textbf{plush beach chair} with its paws on the cushion, set against a background of a \textbf{vibrant sandy beach, choppy waves, and lush tropical scenery...} & ...corgi wearing a colorful straw hat and large sunglasses, sitting on a sunlit beach chair with a \textbf{tropical beach landscape, including palm trees and the ocean waves in the background}. \\
\midrule
\textbf{City Night Scene} & ...lively city street at night with bright lights, towering skyscrapers, and people walking, with \textbf{vibrant colors and realistic lighting effects}, in the background there are \textbf{numerous illuminated signs and decorations}. & ...bustling city street at night with bright lights, tall buildings, and people walking, \textbf{realistic-looking photo with vibrant colors and detailed textures}. & ...lively city street at night with bright lights, tall buildings with illuminated signs, bustling crowds, and \textbf{vibrant city lights surrounding it}, \textbf{realistic photo-like scene with warm and inviting glow}. \\
\midrule
\textbf{Mountain View} & ...stunning mountain landscape with snow-capped peaks, vibrant pine trees, and a clear blue sky, with \textbf{stunning lighting and vibrant colors}. & ...stunning mountain landscape with snow-capped peaks, vibrant pine trees, a clear blue sky \textbf{with fluffy clouds}, realistic photo, \textbf{warm sunset lighting}, \textbf{beautiful natural scenery}. & ...stunning mountain landscape with snow-capped peaks, vibrant pine trees, and a clear blue sky in the background, with \textbf{colorful lighting effects} and a \textbf{fluffy cloud in the sky}. \\
\bottomrule
\end{tabular}
}
\end{table}

\noindent
\begin{minipage}{0.44\textwidth}
    \centering
\includegraphics[width=\linewidth]{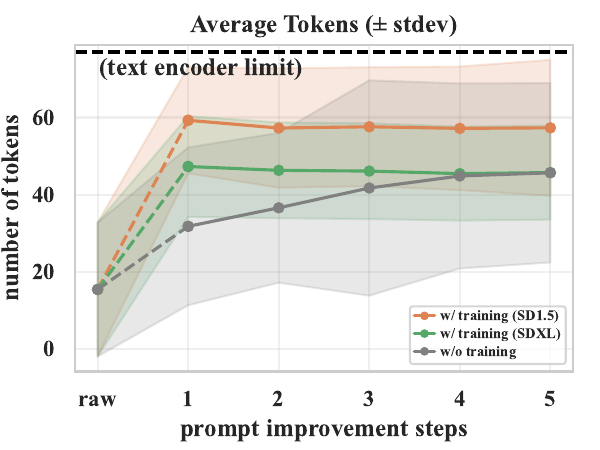}
    \vspace{-1.4em}
    \captionof{figure}{Token counts of raw and refined prompts across improvement steps. Prompt length increases gradually without catastrophic growth or exceeding the text encoder limit (ImageReward task).}
    \label{fig:result-prompt-len}
\end{minipage}
\hfill
\begin{minipage}{0.54\textwidth}
    \centering
\includegraphics[width=\linewidth]{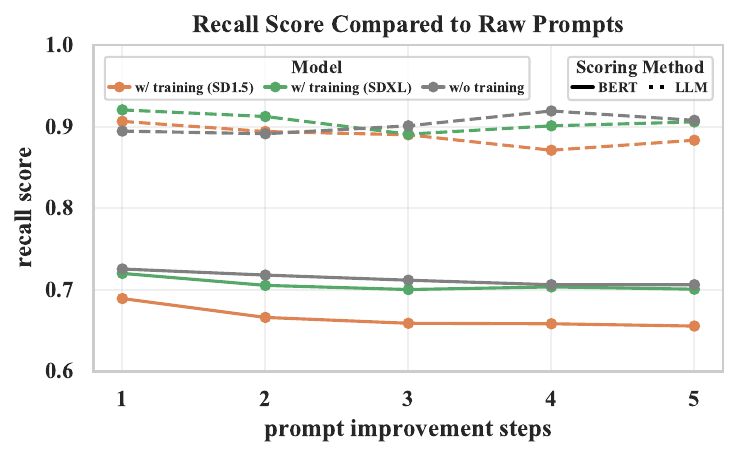}
    \captionof{figure}{Quantitative analysis of semantic drift during prompt refinement. Overall semantic similarity (BERTScore-recall) shows minor changes, while core user intent (LLM-based recall) is well preserved (1.0 indicates identical semantics; ImageReward task).}
    \label{fig:result-prompt-drift}
\end{minipage}

\subsection{Prompt Length Growth}

We further analyze how prompt length evolves under iterative refinement. In our framework, the prompt serves as the sole control signal for the environment (\ie, the diffusion model). Without explicit regularization on prompt length, the policy-refined prompts may continuously grow or exceed the text encoder limit (77 tokens in our setting~\cite{radford2021learning}). Fig.~\ref{fig:result-prompt-len} illustrates the change in prompt length (measured in the number of tokens) across prompt improvement steps under a single-reward setting, where no explicit constraint on prompt length is imposed. As expected, the prompt length increases over time since it is the primary control signal. However, reinforcement learning does not lead to catastrophic length inflation nor does it exceed the text encoder limit. Furthermore, we observe that more text-aligned environments (\eg, SDXL) require fewer tokens for precise control, whereas less aligned environments (\eg, SD1.5) tend to rely on longer prompts. This suggests that the growth in prompt length is driven by the capability of the environment rather than instability in the reinforcement learning process.

\subsection{Semantic Drift in Iterative Prompt Refinement}

Iterative prompt refinement may introduce semantic drift, potentially causing the refined prompt to deviate from the original user intent. However, this effect is mitigated by two key design choices in our framework. First, the policy model has access not only to the prompt from the previous step but also to the original prompt at every refinement step, enabling it to preserve the initial intent. Second, the entire refinement process is treated as a single episode in reinforcement learning, where only the final reward is provided. As a result, any deviation from the original intent leads to negative feedback during training, discouraging semantic drift.

To quantitatively analyze this effect, Fig.~\ref{fig:result-prompt-drift} compares refined prompts with the original prompts using two recall-based similarity metrics: BERTScore-recall~\cite{bert-score} (deberta-xlarge-mnli~\cite{he2021deberta}) for overall semantic similarity, and an LLM-based recall metric\footnote{An LLM (Qwen2.5-VL-3B-Instruct) extracts core user-intent concepts from the raw prompt, and we measure the ratio of concepts retained in the refined prompt.} for core user intent preservation. The results indicate limited semantic change and no significant loss of core intent compared to non-RL prompt refinement, suggesting that reinforcement learning does not exacerbate semantic drift. Consistent with prior observations, more text-aligned environments (\eg, SDXL) exhibit smaller variations, whereas less aligned models (\eg, SD1.5) require larger token-level modifications.

\section{Additional Results} \label{supp-results}

\subsection{Generalization to Recent Flow-matching Models}
\label{supp-flow-models}

To evaluate generalization beyond diffusion models, we assess our trained policy on recent flow-matching text-to-image models that were not seen during training. The policy is trained in a diffusion-model environment (SD1.5 or SDXL with ImageReward) and directly applied to flow-matching model environments without additional fine-tuning. Specifically, we evaluate on SD3.5-large (8B)~\cite{esser2024scaling} and FLUX.1-dev (12B)~\cite{flux1dev2024}. Quantitative results in Tab.~\ref{tab:result-flow-model} demonstrate that our method generalizes effectively to these unseen environments, despite substantial differences in model scale, architecture, training paradigm, and release date.

Since flow-matching models employ a different noise parameterization and scheduler from diffusion models, we align their timesteps by matching the noise-to-signal ratio (SNR). Concretely, each flow-matching timestep is converted into a simulated diffusion timestep with an equivalent SNR, which is then provided as input to the policy model. For sampling, we use the default hyperparameters of each flow-matching model. All PromptLoop sampling hyperparameters are kept identical to those in the main experiments, including the sparse refinement strategy applied at evenly spaced intervals in the original timestep domain of the flow-matching models.

\begin{table}[]
    \centering
\resizebox{0.65\columnwidth}{!}{
\begin{tabular}{clcccc}
\toprule
\textbf{Training Setup} & \textbf{Method} & \textbf{Image Reward} & \textbf{HPSv2} & \textbf{Aesthetics} & \makecell{\textbf{MLLM Score}} \\
\midrule
  \multirow{4}{*}{\shortstack{SDXL \\ \& Image \\ Reward}} & FLUX.1-dev  & 1.001 & \textbf{0.286} & 6.202 & 0.741 \\
 & \textbf{+ ours}  & \textbf{1.246} & \textbf{0.286} & \textbf{6.570} & \textbf{0.757}  \\ \cmidrule(l){2-6}
 & SD3.5-large & 1.079 & \textbf{0.288} & 5.957 & 0.729 \\
 & \textbf{+ ours} & \textbf{1.254} & \textbf{0.288} & \textbf{6.197} & \textbf{0.744} \\ \midrule
 \multirow{4}{*}{\shortstack{SD1.5 \\ \& Image \\ Reward}} & FLUX.1-dev & 1.001 & \textbf{0.286} & 6.202 & 0.741 \\
 & \textbf{+ ours} & \textbf{1.258} & \textbf{0.286} & \textbf{6.542} & \textbf{0.758} \\ \cmidrule(l){2-6}
 & SD3.5-large & 1.079 & \textbf{0.288} & 5.957 & 0.729 \\
 & \textbf{+ ours} & \textbf{1.242} & 0.287 & \textbf{6.229} & \textbf{0.744} \\
\bottomrule
            \end{tabular}}
    \caption{Quantitative results demonstrating zero-shot generalization to recent flow-matching models.}
    \label{tab:result-flow-model}
\end{table}

\subsection{Denoised Estimate as an MLLM input}
\label{supp-tweedie}

\begin{table*}[!t]
\begin{minipage}[t]{1.00\linewidth}
    \centering
    \includegraphics[width=0.80\linewidth]{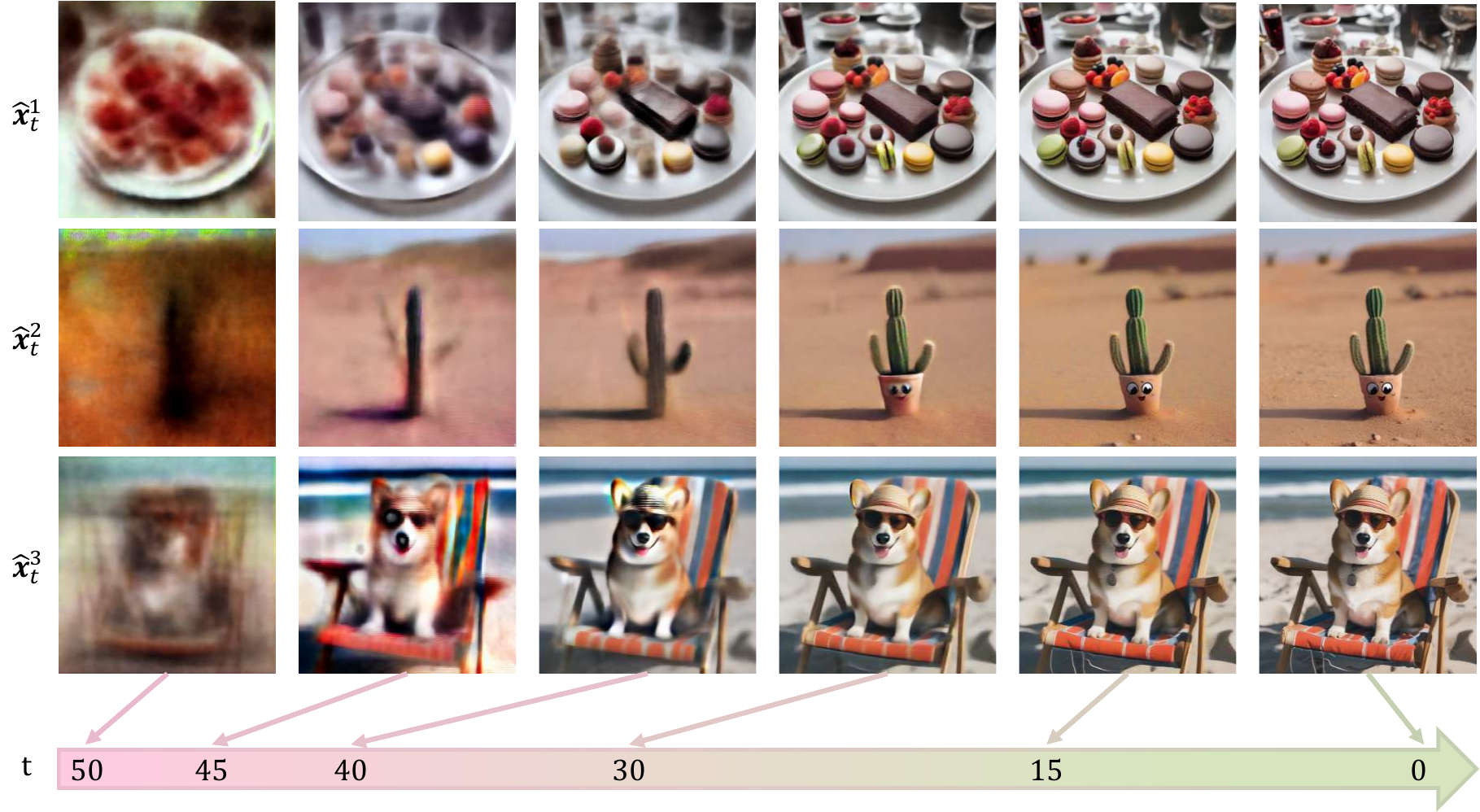}
    \vspace{1.0em}
\end{minipage} 
\begin{minipage}[t]{1.00\linewidth}
    \centering
    \resizebox{0.80\textwidth}{!}{
    \begin{tabular}{lcp{0.75\textwidth}}
    \toprule
     \textbf{Sample} & \textbf{$t$} & \textbf{MLLM Generated Description} \\ \midrule
     0 & 50 & The image is blurry and shows \textbf{a plate with red berries}, possibly raspberries or strawberries, on \textbf{a white plate}. \\
     & 45 & The image is blurry and shows \textbf{a plate with various food items}, including what appears to be fruits and possibly some vegetables or grains. \\
     & 40 & The image is a blurry photograph of \textbf{a plate with various pastries and desserts}, including \textbf{macarons and other confections}. \\ \midrule
     1 & 50 & The image appears to be a blurred, abstract representation of a \textbf{landscape with warm colors},
     possibly featuring a body of water and some vegetation or terrain. \\
     & 45 & The image appears to be a blurred photograph of a person's face, with the focus on the nose and mouth area. The background is indistinct, but it seems to be \textbf{an outdoor setting} with some greenery visible. \\
     & 40 & The image depicts \textbf{a cactus in a desert landscape} with \textbf{reddish-brown sand} and a \textbf{clear sky}. The cactus has \textbf{a long, thin trunk} with \textbf{two small branches} extending from it. The overall scene is arid and typical of a desert environment. \\ \midrule
     2 & 50 & The image appears to be a blurred photograph of a person \textbf{wearing a hat} and a light-colored top, \textbf{possibly outdoors} with greenery in the background. \\
     & 45 & The image depicts \textbf{a dog} holding a stack of books, with a \textbf{colorful background.} The dog appears to be \textbf{wearing sunglasses} and is standing on a table or surface. \\
     & 40 & The image shows \textbf{a cartoon dog wearing sunglasses and a hat}, \textbf{sitting on a striped beach chair by the ocean}. The dog appears relaxed and is enjoying a sunny day at the beach. \\
    \bottomrule
    \end{tabular}
    }
    \captionof{figure}{
        \textbf{Top: Visualization of the denoised estimates along the diffusion sampling trajectory. Bottom: Descriptions generated by the MLLM policy model conditioned on these denoised estimates.} Denoised estimates provide identifiable visual states even at early diffusion timesteps, for both humans and the MLLM policy model.
    }
    \vspace{-1.0em}
    \label{fig:tweedie}
  \end{minipage}
\end{table*}

We feed the visual state $\bm{x}_t$ to the MLLM policy model in the form of its denoised estimate $\hat{\bm{x}}_t$. Using $\hat{\bm{x}}_t$ as an approximation of the final sample $\bm{x}_0$ is a well-established practice in the diffusion literature~\cite{yu2023freedom, chung2022diffusion, kim2025free2guide, singhal2025general}, regardless of the specific guidance mechanism, energy function formulation, or evaluation function used. For example, denoised estimates are commonly used as inputs to a wide range of energy functions, including measurement-consistency objectives~\cite{chung2022improving, chung2022diffusion}, neural network–based energy models~\cite{yu2023freedom}, and MLLM reward models~\cite{kim2025free2guide, singhal2025general}.

As shown in Fig.~\ref{fig:tweedie}, the denoised estimate $\hat{\bm{x}}_t$ contains semantically meaningful information even at early diffusion timesteps, despite the inherent blur introduced by its expectation-based formulation. Importantly, these blurred estimates remain interpretable to humans, as coarse object shapes, colors, and global layouts are still recognizable even though fine details are missing. The descriptions generated by the MLLM policy model exhibit a similar behavior. The MLLM can analyze and explain these early denoised estimates in much the same way humans do when presented with a blurred image. This empirical evidence indicates that denoised estimates provide identifiable visual states for both humans and the MLLM policy model, which enables them to generate meaningful guidance throughout the diffusion sampling trajectory. In addition, prior work shows that the early phase of diffusion sampling primarily captures the low-frequency structure of the image, such as global shapes and coarse layout~\citep{yu2023freedom}. At this stage, high-level textual guidance is especially relevant, and the blurred $\hat{\bm{x}}_t$ is sufficiently informative to support such guidance.

\subsection{Training Dynamics and Stability}

\begin{table}
\centering
\begin{minipage}[t]{0.54\columnwidth}
\centering
\captionof{figure}{Training dynamics of proposed framework, showing stable optimization. Stronger policy initialization and improved environment lead to faster convergence and more effective reward alignment (ImageReward).}
\label{fig:wandb}
\includegraphics[width=\columnwidth]{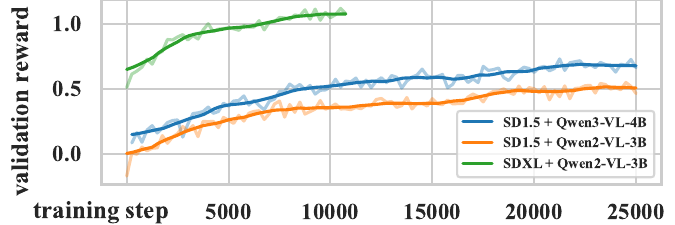}
\end{minipage}
\hfill
\begin{minipage}[t]{0.45\columnwidth}
\centering
\caption{Ablation results showing that stronger policy initialization improves reward alignment (SD1.5 \& ImageReward).}
\label{tab:abl-policy}
\vspace{2.5em}
\resizebox{\columnwidth}{!}{
\begin{tabular}{lcccc}
\toprule
\makecell[l]{\textbf{Policy}\\\textbf{Model}} & \makecell{\textbf{Image}\\\textbf{Reward}} & \makecell{\textbf{HPS}\\\textbf{v2}} & \textbf{Aesthetics} & \makecell{\textbf{MLLM}\\\textbf{Score}} \\
\midrule
   Qwen2.5-VL-3B & 0.6320 & 0.2701 & 5.853 & 0.725 \\
   Qwen3-VL-4B & \textbf{0.6922} & \textbf{0.2705} & \textbf{6.024} & \textbf{0.730} \\
\bottomrule
\end{tabular}}
\end{minipage}
\end{table}

Despite employing RL, our training remains stable, as shown in Fig.~\ref{fig:wandb}. This stability primarily stems from three factors: (i) initialization from a strong MLLM-based policy, (ii) parameter-efficient updates via LoRA, and (iii) a fixed diffusion environment that interacts with the policy only through prompts, reducing non-stationarity. Combined with GRPO, these design choices effectively mitigate common RL instabilities such as oscillatory behavior.

Fig.~\ref{fig:wandb} further reveals that both convergence speed and final reward are largely determined by the quality of the policy initialization and the environment, rather than the intrinsic instability of RL itself. In particular, stronger initial policies lead to faster and more stable optimization trajectories. This trend is quantitatively supported in Tab.~\ref{tab:abl-policy}, where replacing the policy with a more capable model consistently improves all evaluation metrics. These results highlight that policy initialization is a critical factor for achieving reliable and efficient preference alignment.

\subsection{More Qualitative Samples} \label{supp-single-rewards}

\begin{figure}[t]
\begin{center}
\includegraphics[width=1.0\textwidth]{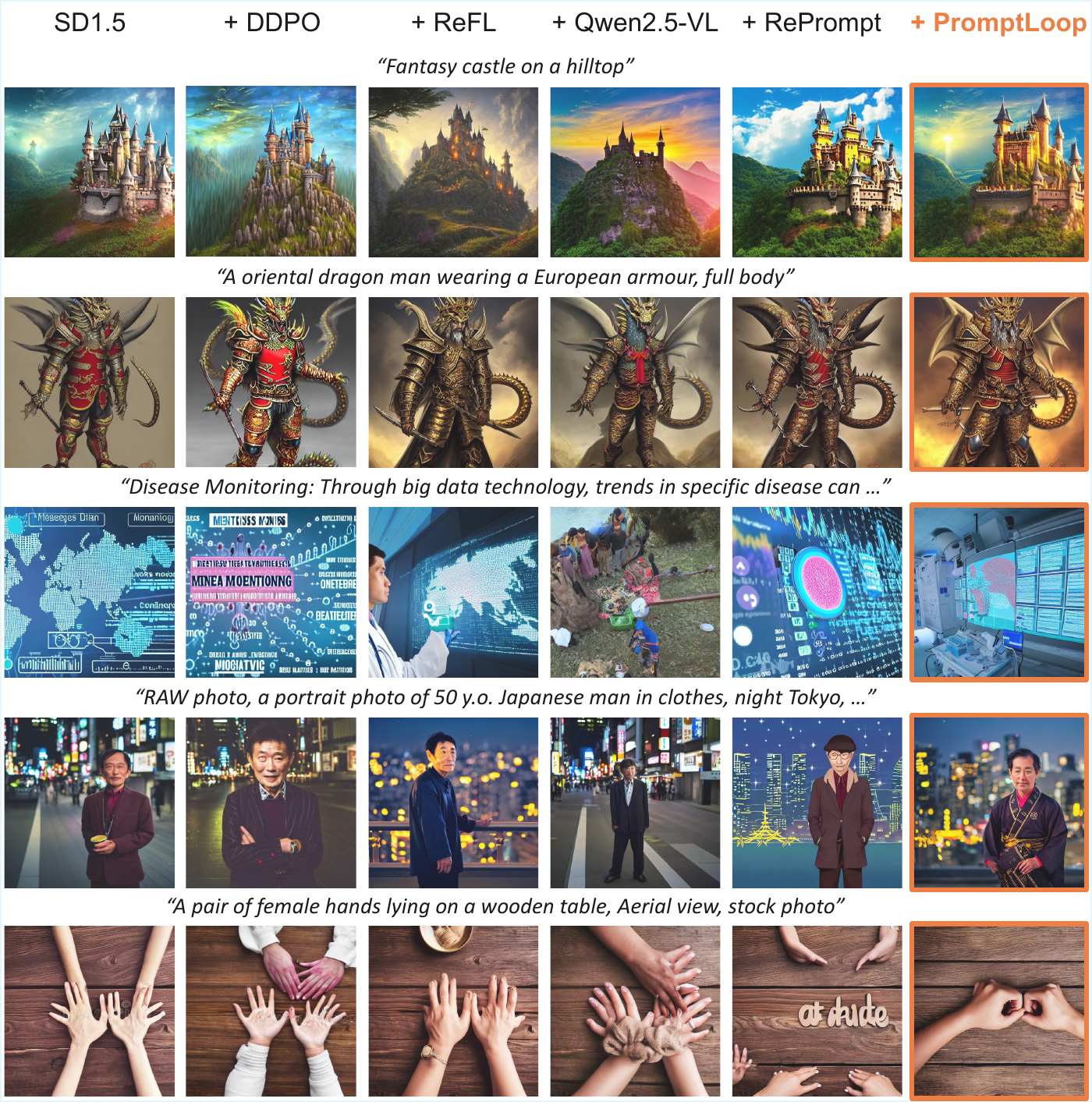}
\end{center}
\vspace{-1em}
\caption{Qualitative comparison of baseline methods (SD1.5 \& ImageReward).}
\label{fig:result-ir-compare}
\end{figure}

\begin{figure}[t]
\begin{center}
\includegraphics[width=1.0\textwidth]{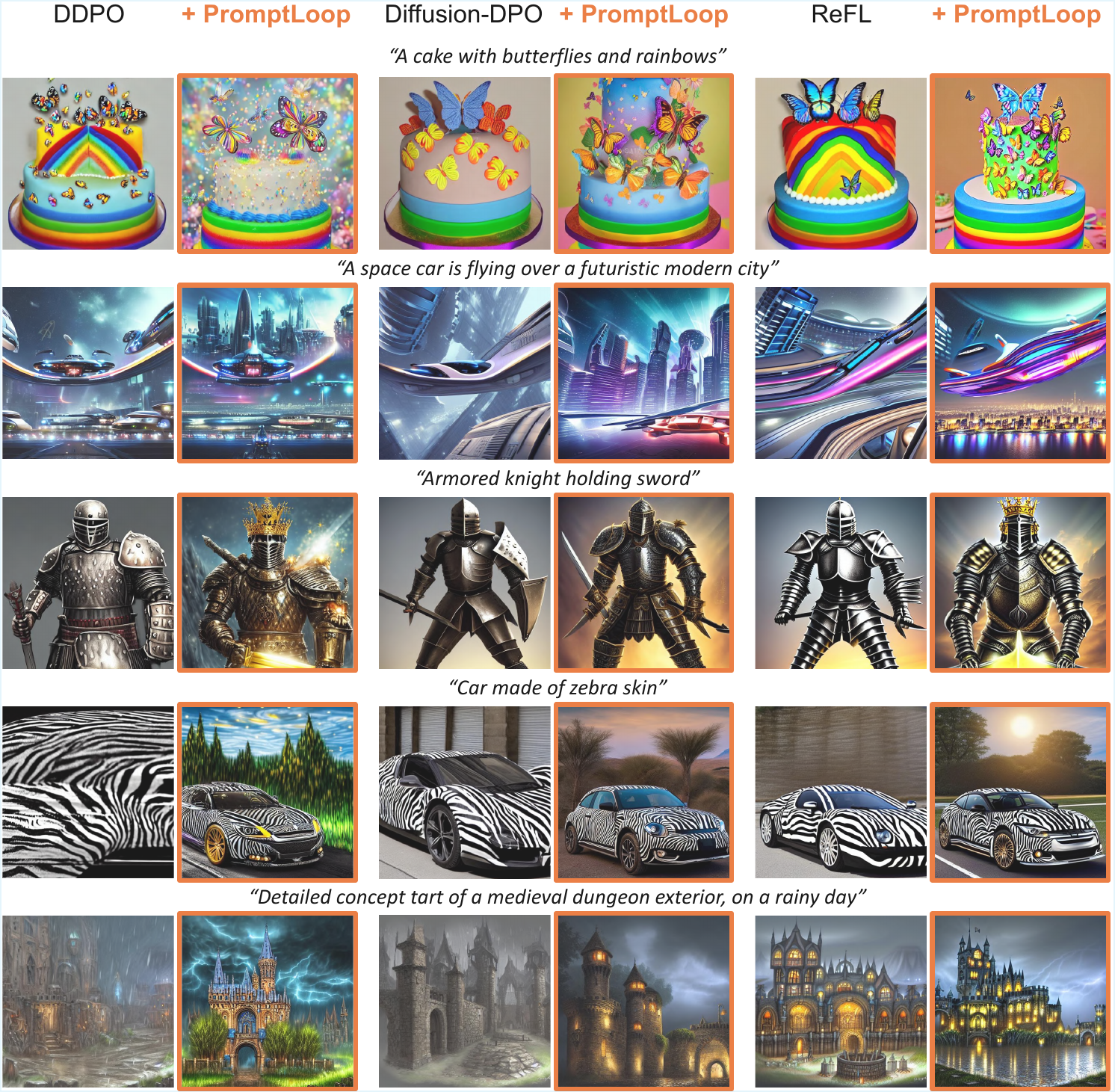}
\end{center}
\vspace{-1em}
\caption{Qualitative results demonstrating the orthogonality of our method compared with reward-aligned baselines (SD1.5 \& ImageReward).}
\label{fig:result-ir-ortho}
\end{figure}

\begin{figure}
\vspace{-3em}
\begin{center}
\includegraphics[width=0.8\textwidth]{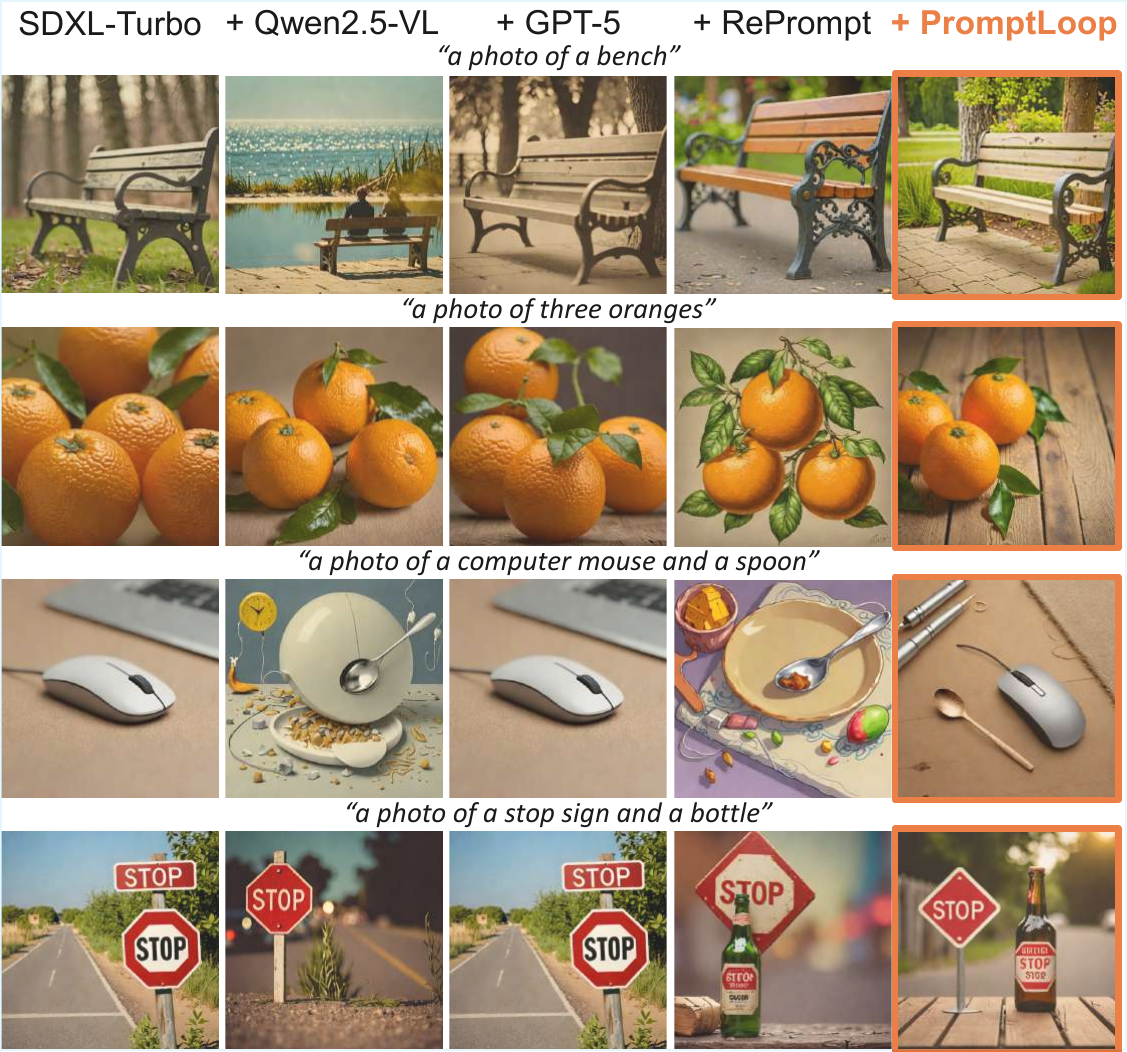}
\end{center}
\vspace{-1em}
\caption{Qualitative comparison of composite-reward alignment, illustrating improvements over baseline methods. (SDXL-turbo \& RePrompt)}
\label{fig:result-rp-compare}
\vspace{-3em}
\end{figure}

\begin{figure}
\begin{center}
\includegraphics[width=0.65\textwidth]{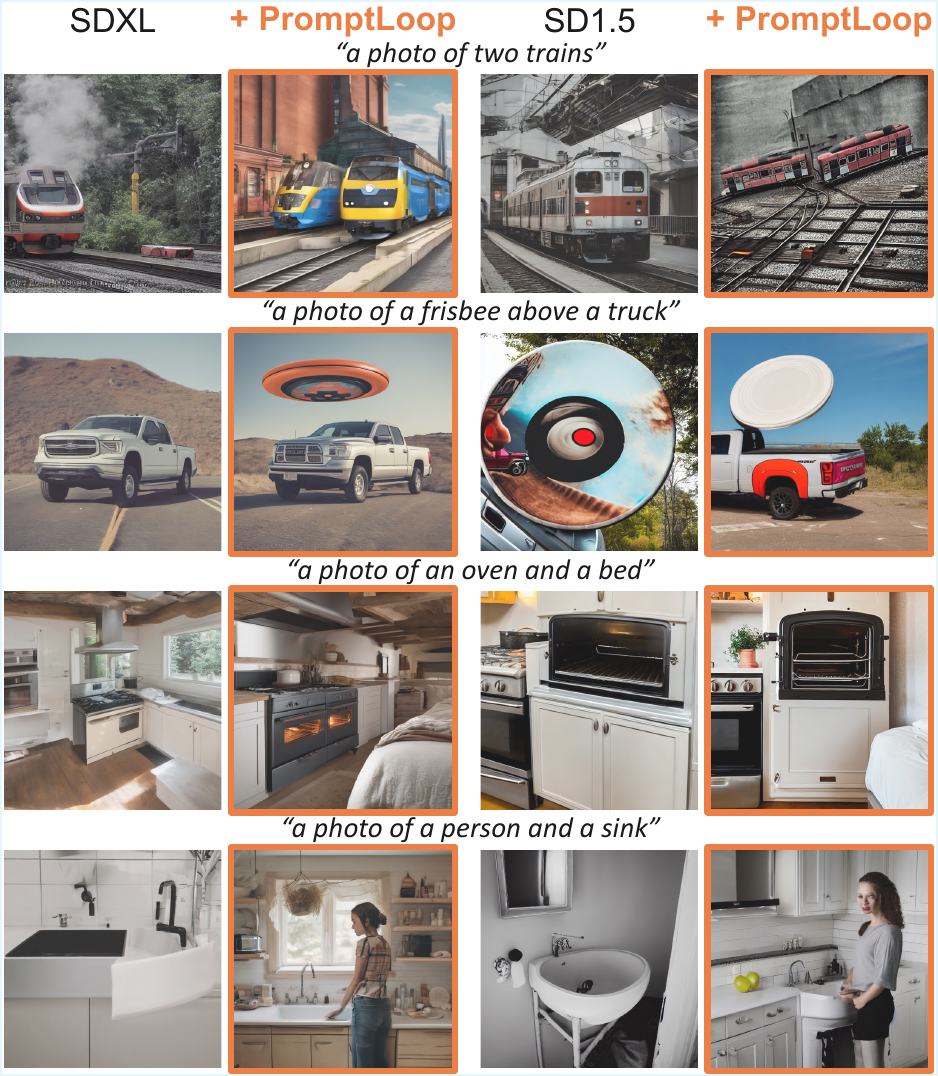}
\end{center}
\vspace{-0.5em}
\caption{Qualitative results showing the orthogonality and generalizability achieved by applying our method to unseen reward-alignment baselines (SDXL-turbo \& RePrompt).}
\label{fig:result-rp-ortho}
\end{figure}

\begin{figure}
  \centering
  \includegraphics[width=1.0\textwidth]{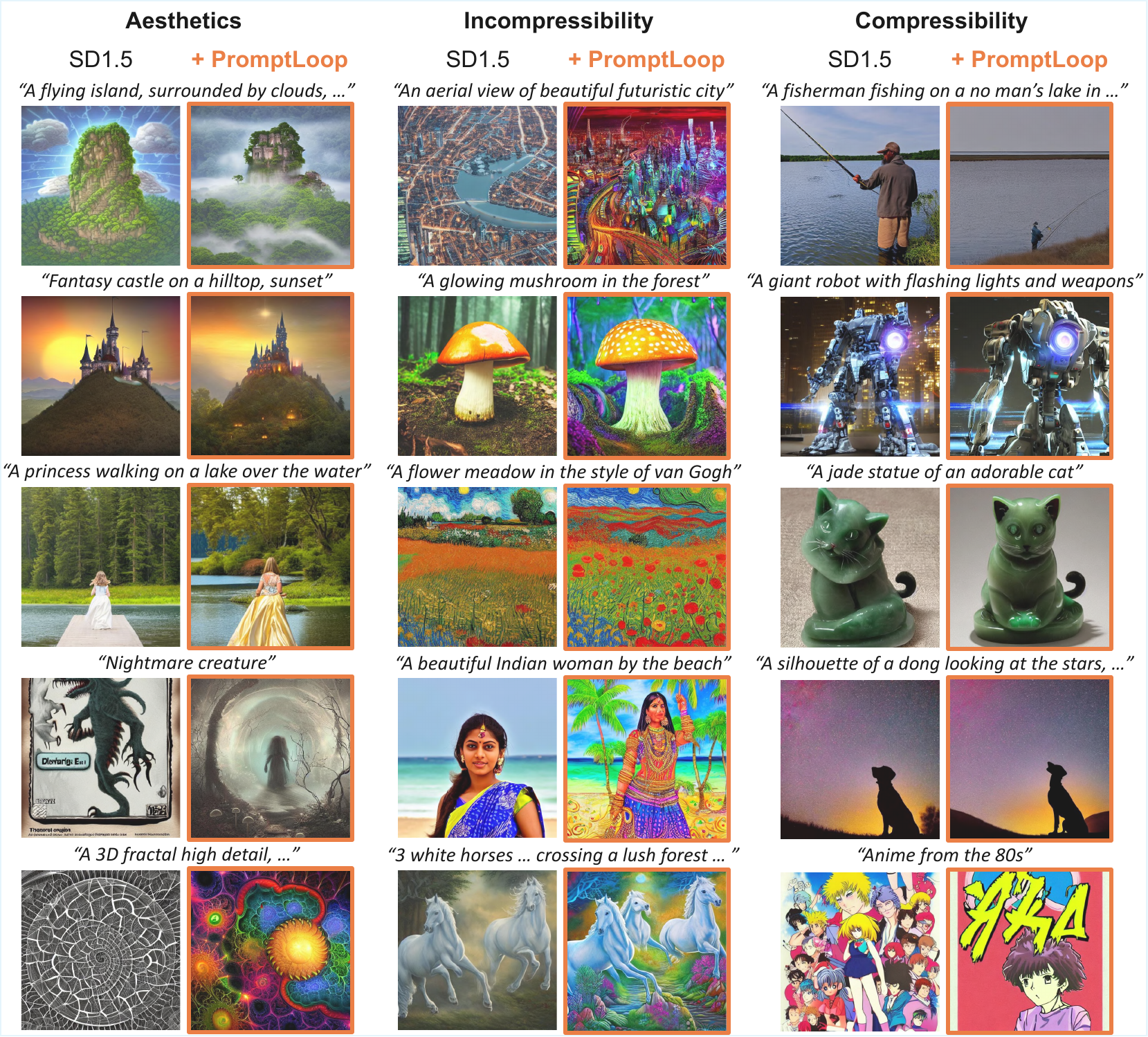}
  \caption{Qualitative results demonstrating the applicability of our framework to diverse reward signals.}
  \label{fig:result-extra}
\end{figure}

We present qualitative samples corresponding to the quantitative evaluation of single-reward alignment on SD1.5 and composite-reward alignment on SDXL-turbo reported in Tab.~\ref{tab:sigle-reward} and Tab.~\ref{tab:complex-reward}, which could not be included in the main text due to space constraints. Specifically, Fig.~\ref{fig:result-ir-compare},~\ref{fig:result-rp-compare} illustrates comparisons against baseline reward alignment methods, Fig.~\ref{fig:result-ir-ortho},~\ref{fig:result-rp-ortho} highlights the orthogonality of our approach to other reward alignment techniques. The results, consistent with the quantitative findings, demonstrate clear advantages in prompt alignment and human preference, while also highlighting the orthogonality and generalization capability of our approach. In addition to ImageReward as a single-reward task, we also trained models using aesthetic quality~\citep{schuhmann_improved_aesthetic_predictor_2025}, compressibility, and incompressibility rewards~\citep{black2024training}, as shown in Fig.~\ref{fig:result-extra}. These experiments further demonstrate that our proposed framework can be generally applied across diverse reward types.

\section{LLM Usage}

Large Language Models (LLMs) were used solely as an editorial aid to improve the clarity and readability of the manuscript. Specifically, LLMs assisted in polishing grammar, refining sentence structure, and ensuring consistency in style. They were not used in any aspect of research ideation, experimental design, data analysis, or in the generation of substantive scientific content. All ideas, results, and interpretations presented in this paper are the responsibility of the authors.

\end{document}